\documentclass[lettersize,journal]{IEEEtran}
\usepackage{amsmath,amsfonts}
\usepackage{algorithm}
\usepackage{array}
\usepackage[caption=false,font=normalsize,labelfont=sf,textfont=sf]{subfig}
\usepackage{textcomp}
\usepackage{stfloats}
\usepackage{url}
\usepackage{verbatim}
\usepackage{graphicx}
\usepackage{cite}
\usepackage{color}

\usepackage{mathtools}
\usepackage[only,llbracket,rrbracket]{stmaryrd}

\usepackage{makecell}
\usepackage{xtab,afterpage}
\usepackage{tabularray}
\usepackage{amsmath}
\usepackage{verbatim} 
\usepackage{lipsum}
\usepackage{multirow}
\usepackage{rotating}
\usepackage{placeins}

\usepackage{algpseudocode}
\usepackage[numbers]{natbib}

\newcommand{\Cai}[1]{\textcolor{blue}{\textbf{/* #1 (Cai) */}}}

\begin{document}

\title{Exploring Causal Learning through Graph Neural Networks: An In-depth Review}

\author{Simi~Job, 
        Xiaohui~Tao, 
        Taotao Cai,
        Haoran Xie, 
        Lin Li,
        Jianming Yong
        and~Qing Li
        


\thanks{*Corresponding authors: Simi Job (email: \textit{simi.job@unisq.edu.au}) and Xiaohui Tao (email: \textit{Xiaohui.Tao@unisq.edu.au}) are with School of Mathematics, Physics, and Computing, University of Southern Queensland, Australia}
\thanks{Taotao Cai is with School of Mathematics, Physics, and Computing, University of Southern Queensland, Australia (e-mail: \textit{Taotao.Cai@unisq.edu.au}).}
\thanks{Haoran Xie is with the Department of Computing and Decision Sciences, Lingnan University, Tuen Mun, Hong Kong (e-mail: \textit{hrxie@ln.edu.hk})}
\thanks{Lin Li is with the School of Computer and Artificial Intelligence, Wuhan University of Technology, China (e-mail: \textit{cathylilin@whut.edu.cn})}
\thanks{Qing Li is with the Department of Computing, Hong Kong Polytechnic University, Hong Kong Special Administrative Region of China (e-mail: \textit{qing-prof.li@polyu.edu.hk})}
\thanks{Yong is with the School of Business, University of Southern Queensland, Springfield (e-mail: jianming.yong@unisq.edu.au)}
}


\markboth{Journal of \LaTeX\ Class Files,~Vol.~14, No.~8, August~2021}%
{Shell \MakeLowercase{\textit{et al.}}: A Sample Article Using IEEEtran.cls for IEEE Journals}


\maketitle

\begin{abstract}
In machine learning, exploring data correlations to predict outcomes is a fundamental task. Recognizing causal relationships embedded within data is pivotal for a comprehensive understanding of system dynamics, the significance of which is paramount in data-driven decision-making processes. Beyond traditional methods, there has been a surge in the use of graph neural networks (GNNs) for causal learning, given their capabilities as universal data approximators. Thus, a thorough review of the advancements in causal learning using GNNs is both relevant and timely. To structure this review, we introduce a novel taxonomy that encompasses various state-of-the-art GNN methods employed in studying causality. GNNs are further categorised based on their applications in the causality domain. We further provide an exhaustive compilation of datasets integral to causal learning with GNNs to serve as a resource for practical study. This review also touches upon the application of causal learning across diverse sectors. We conclude the review with insights into potential challenges and promising avenues for future exploration in this rapidly evolving field of machine learning.
\end{abstract}



\section{Introduction}\label{sec:introduction}

Exploration of causality is a foundation of machine learning, aiming to uncover the intricate relationships embedded within data features. Predominantly, the focus has been on identifying correlations and associations, such as the links between lifestyle choices and health outcomes. Though traditional machine learning techniques such as classification are well-suited to predicting outcomes, they are inadequate for understanding causality. Moreover, there has been a growing recognition of the need to delve deeper into the actual causal elements behind these associations, especially when the objective extends beyond mere prediction to actual improvement or intervention. In building on this approach, it is essential to determine an alternative course of action that can lead to a more optimal outcome inclusive of cause-effect relationships, which is where the concept of causality is significant. The process of causal analysis can be adopted to explore such causal effects, whereby the elements that contribute to a specific outcome can be determined. 

Causal analysis is crucial in machine learning to ensure generalizability, explainability and fairness of ML models \cite{zhao2022machine}. At its core, causality bifurcates into causal discovery and causal inference. The former unravels the structural relationships within data, while the latter delves into the ramifications of different interventions. A significant challenge in causal inference is discerning the distinct effects of concurrent actions. A key deterrent to determining causation is the assumption that causal effects cannot be reliably evaluated statistically \cite{alma991006437348204691}, on account of the impracticality of knowing possible outcomes in the event of an alternative course of action.

Causality learning has gained significance in several areas, including in the development of recommender systems \cite{si2023enhancing}, language modeling \cite{vig2020investigating}, the medical field \cite{chai2020diagnosis} (including medical imaging \cite{castro2020causality}) and pharmacology \cite{zhao2022machine}, urban intelligence \cite{mannering2020big}, economics \cite{alam2020causal}, and so on. In causal analysis, it is essential to address counterfactuals that allow replication of different approaches simultaneously. Causal machine learning provides a framework for including this aspect of causality and discovers the influence of confounders on an outcome. Incorporating causality into machine learning establishes causal relationships between features, which consequently leads to improved model accuracy \cite{carbo2020machine}. Causal forests, a variation of random forests (RF) \cite{carbo2020machine} and ensemble methods that use logistic regression, RF, gradient boosting and SVM \cite{decruyenaere2020obesity} are examples of methods employed in causal machine learning. Propensity score, covariate balancing, instrumental variable (IV) and frontdoor criterion are a few conventional methods adopted for learning causal effects. Traditional causal discovery methods include constraint-based (PC, FCI) and score-based (GES) methods. Fundamentally, causal interactions can be represented in the form of graphs in most cases, rendering such traditional approaches ineffective for handling the dynamics of causal relationships. 

\begin{figure}
    \centering
    \includegraphics{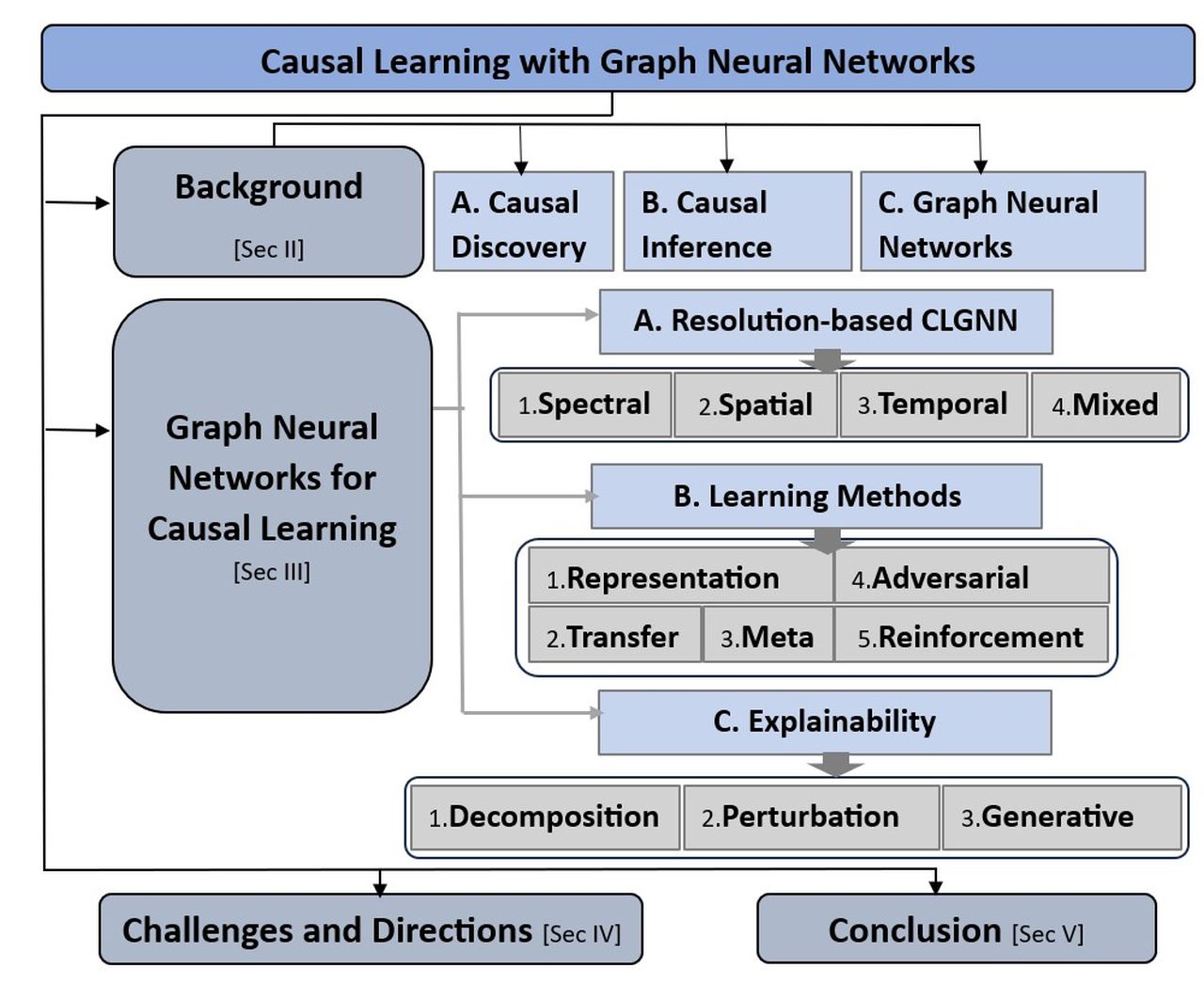}
    \caption{Survey Overview}
    \label{fig:survey}
\end{figure}

Deep learning techniques such as MLPs (multilayer perceptron) have been used to learn causality \cite{si2023enhancing}, with graph-based deep learning methods adopted in recent research studies \cite{chai2020diagnosis}. Modeling causal relationships by means of graphs serve to establish causality dynamics. Causality learning using graph methods are gaining attention in recent times, since GNNs have proven to be effective for causal analysis due to their potential as universal approximators of data \cite{zevcevic2021relating}. Moreover, GNNs are adept at multi-modal deep learning, and can also address the problem of feature sparsity through graph representations by means of feature interactions represented in the form of nodes \cite{zhai2023causality}. GNN, with its NN architecture, can handle complex graphs consisting of a multitude of nodes. Moreover, GNNs are capable of extracting information from unstructured data, as well as for modeling dynamic and evolving data structures. On these grounds, GNNs have an inherent potential to capture causality in data, especially when compared to traditional approaches as discussed in \cite{feng2021should,zhao2022learning}.

Existing surveys on graphical causal learning predominantly centered on generalized graphical models. While \cite{cheng2022data} delved into graphical causal modeling techniques, it overlooked the role of GNNs. Other survey papers like \cite{wu2020comprehensive} and \cite{zhou2020graph} have explored GNNs, but disregarded their applications in causal learning. \cite{yuan2022explainability} focused on the explainability aspect of GNNs, with a brief reference to the causal screening method. Notably, causality-focused surveys such as \cite{nogueira2022methods} and \cite{yao2021survey} omitted GNNs for causality, while \cite{kuang2020causal} provided only a cursory mention of GNNs in their discourse on causal inference.

This survey comprehensively reviews GNNs for causal learning, addressing this identified gap in the literature. To our knowledge, this is the only study focusing on causal learning with GNNs. The key contributions of the article are the following:

\begin{itemize}
    \item \textbf{Systematic Taxonomy and In-depth Review:} We introduce a meticulously designed taxonomy tailored for this survey, showcased in Fig.~\ref{fig:survey}. Every category undergoes a thorough review, summarization, and comparison against key works in the domain. Furthermore, we compile a detailed list of state-of-the-art methods employed in causal learning with GNNs, along with datasets pertinent to causality-focused GNN research (Sec.~\ref{sec-CGN}).
    
    \item \textbf{Rich Resources and Noteworthy References:} We present a curated collection of open resources, which encompass benchmark datasets and technical details. Our cited references are drawn from premier peer-reviewed journals covering areas such as data mining, AI, GNNs, and knowledge discovery (Sec.~\ref{sec-CGN}). 
    
    \item \textbf{Future Horizons:} We recognize the evolving nature of GNNs and its application in causality. Using this knowledge, we delve into its present constraints, pressing challenges, and untapped opportunities, charting out potential avenues for future exploration in the field, and offer insights about future research in the domain (Sec.~\ref{sec-CD}).
\end{itemize}

The remainder of the paper is organized as follows: Section~\ref{sec-CLGNN} introduces causal learning concepts and GNNs, including their common variants. Section~\ref{sec-CGN} presents the GNN approaches for causal learning, then Section~\ref{sec-CD} discusses challenges and future directions, and Section~\ref{sec-CON} concludes this survey.


\section{Background}\label{sec-CLGNN}

\begin{figure}
    \centering
    \includegraphics{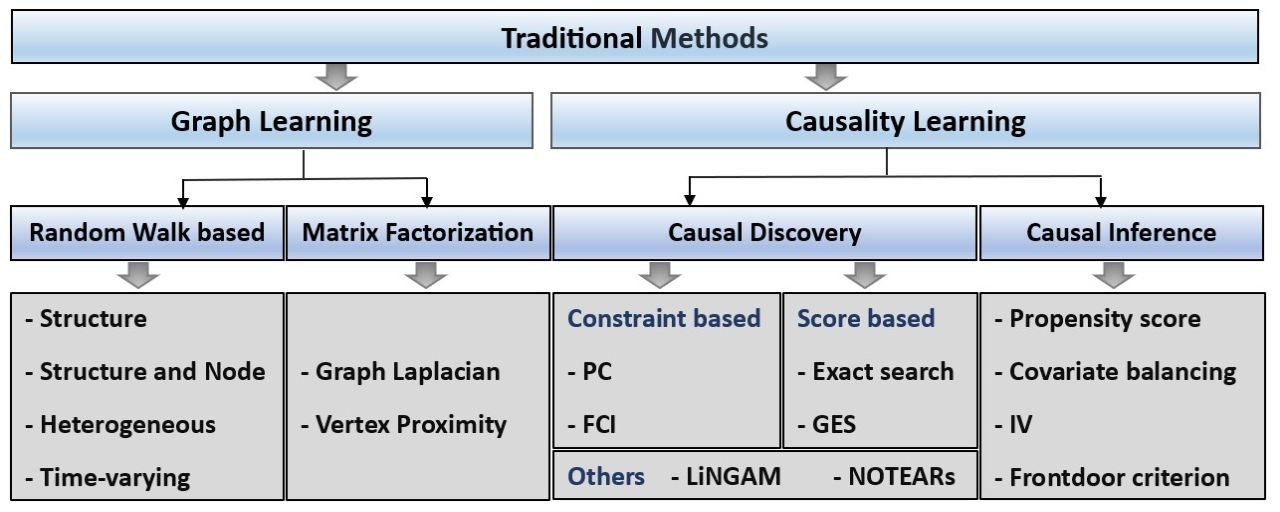}
    \caption{Traditional graph learning and causal learning methods}
    \label{fig:Trad}
\end{figure}

Prior to deep learning methods, causal learning was studied using traditional methods. Fig.~\ref{fig:Trad} provides an outline of traditional methods used in graph learning and causal learning. GNNs, owing to their ability to act as functional approximators, have been researched in recent times due to the potential to extract causal relations and causal effects \cite{yu2019dag}. This section provides a background on causal learning and GNNs.

\subsubsection*{Definition 1 (Structural Causal Model- SCM)}

SCM is a set of endogenous variables $V$ and a set of exogenous variables $U$, mapped in terms of a set of functions $F$. The values of $V = \{V{_0}, V{_1}, ..., V{_i}\}$ and $U = \{U{_0}, U{_1}, ..., U{_u}\}$ are determined by $F$. A SCM is associated with a directed acyclic graph (DAG), with a node in $U$ or $V$, and each edge in $F$.

\subsubsection*{Definition 2 (Causal Modeling)} Causal modelling represents causal relationships mathematically through SCMs. \textit{Potential outcome framework}, similar to the SCM model, is defined in terms of treatment applied to a unit under study. \textit{Potential outcome} is the outcome of a treatment. \textit{Counterfactual outcome} forms the outcome if a different treatment was applied to the unit. Causal learning evaluates the change in outcome when an alternative treatment is applied, where control \textit{c} is altered to treatment \textit{t}, expressed as $\mathbb{E}[y|t] - \mathbb{E}[y|c]$.

\subsubsection*{Definition 3 (Graphs)} A graph  $G$ is a set of nodes with edges and is represented as $G =(V,E)$, where $V$ is the set of vertices (nodes) and $E$ is the set of edges connecting the nodes. Graphs may be directed or undirected. Two nodes connecting an edge $e$ viz. $u \in V$ and $v \in V$ are neighbours and hence have an \textit{adjacent} relationship. Adjacency matrix represents a graph as boolean values for indicating these node connections.
 A \textit{causal graph} is a directed graph defining causal effects between variables. The nodes in a causal graph represent variables such as the treatment, outcome, and other observed or unobserved variables. 

\subsubsection*{Definition 4 (Back-door path)} For a pair of treatment and outcome variables $(t,y)$, a path that connects these two variables is a back-door path for $(t,y)$ if it is not a directed path and it is not blocked.
\subsubsection*{Definition 5 (Confounder)} For a pair of variables $(t,y)$, a variable $z \not\in \{t,y\}$ is a confounder of the causal effect $t \rightarrow y$, if it is the central node of a fork on a back-door path of $(t,y)$.

Causal learning involves causal discovery and causal inference. In causal discovery, the causal structure is derived from data to form a causal graph. Causal inference involves inferring causal effects from this causal graph. These concepts are briefly discussed in the following subsections.

\subsection{Causal Discovery}

 Causal discovery involves analysing observational data to derive causal relations and is based upon four assumptions:
 
 \begin{itemize}
    \item \textit{Acyclicity}: No cycles in the graph and no variable can be a cause of itself
    \item \textit{Markov Property}: A node in a graph is independent of other nodes that are non-descendants, conditional on its direct causes
    \item \textit{Faithfulness}: Causally connected variables in a graph are probabilistically dependent
    \item \textit{Sufficiency}: All confounders in the graph must be observed variables
\end{itemize}

\subsection{Causal Inference}

\begin{figure}
    \centering
    \includegraphics{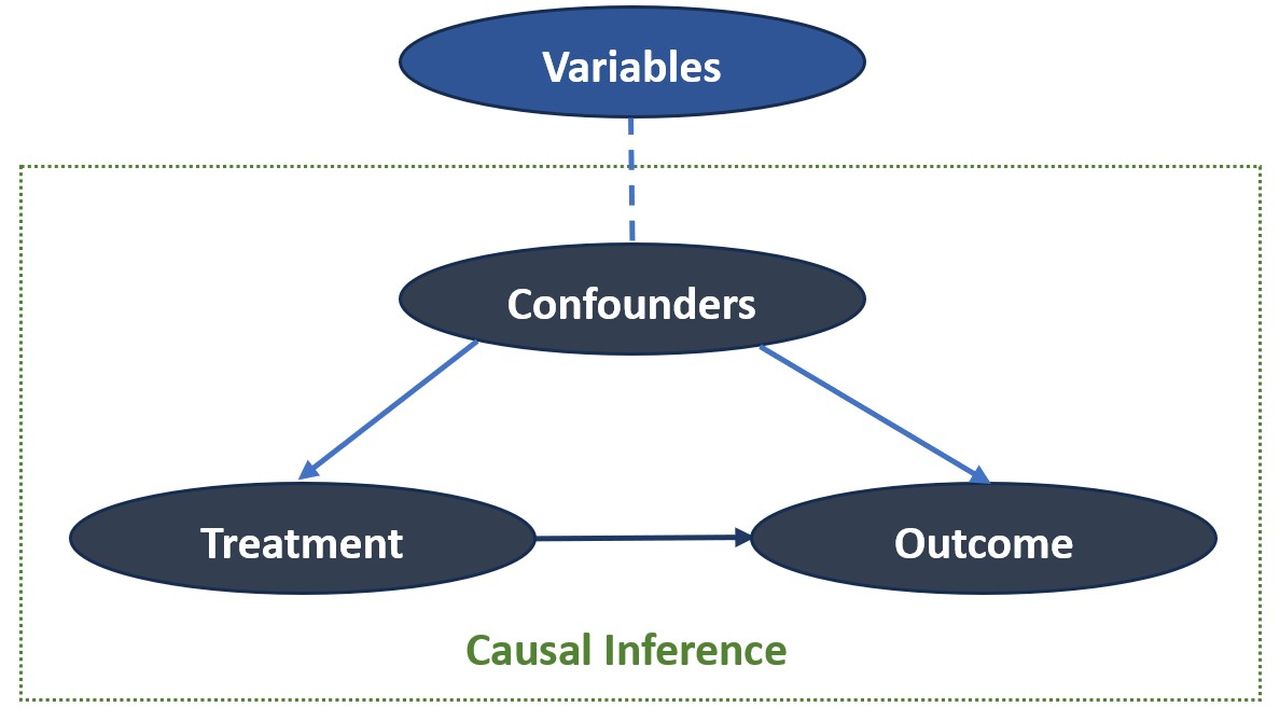}
    \caption{Causal inference}
    \label{fig:CI}
\end{figure}

Causal inference estimates the causal effect of a variable (treatment) over a variable of interest (outcome) as depicted in Fig.~\ref{fig:CI}. Here, a confounder is a variable that is causally associated with treatment and outcome. Causal inference derives the effect of an intervention on the outcome and requires the following conditions to be satisfied:

  \begin{itemize}
    \item \textit{Covariation}: A change in the causal variable should result in a change in the variable that is expected to be affected
    \item \textit{Temporal precedence}: If a causal variable causes an effect on the affected variable, then the occurrence of the causal variable must precede the effect on the affected variable
    \item \textit{Control extraneous variables}: Elimination of possible alternative causes
\end{itemize}
 
Causal inference is based on the following assumptions:
 
 \begin{itemize}
    \item \textit{SUTVA (Stable Unit Treatment Value Assumption)}: No variation in treatment and no interference in the treatment status of a unit
    \item \textit{Ignorability}: Treatment assignment is independent of the outcome
    \item \textit{Consistency}: The potential and the observed outcomes are consistent
    \item \textit{Positivity}: Any observation has a positive chance of getting the treatment
\end{itemize}

In causal learning, two approaches for computing causal effects from observational data are the Backdoor and Frontdoor Adjustment criteria, given a causal graph.

\subsubsection{The Backdoor Adjustment}
 In a causal graph, a set of variables $V$ satisfies the Backdoor criterion for a pair of variables $(X,Y)$ if:
    \begin{itemize}
    \item $V$ blocks all incoming paths into $X$, and
    \item No node in $V$ is a descendant of $X$
    \end{itemize}

\subsubsection{The Frontdoor Adjustment}
 In a causal graph, a set of variables $V$ satisfies the Frontdoor criterion for a pair of variables $(X,Y)$ if:
    \begin{itemize}
    \item $V$ blocks all directed paths from $X$ to $Y$,
    \item All backdoor paths from $V$ to $Y$ are blocked by $X$, and 
    \item There exists no backdoor paths from $X$ to $V$ 
    \end{itemize}

\subsection{Graph Neural Networks}\label{sec-GN}

GNNs are deep learning techniques that are used for processing and analysing graph-based data. With their neural architecture, they can learn node representation. The representation matrix is defined as $H \in R^{N \times F}$, where $F$ is the node representation dimension. GNNs aggregate representations of node neighbours and update node representations in an iterative manner, and are capable of learning causality with their functional approximation properties. A mathematical representation of the GNN framework is defined in Eq.~\ref{eq-GNNeqn} \cite{wu2022graph}, with \textit{Aggregate} and \textit{Combine} functions in each layer, where the node representation is initialized as $H^0 = X$. Here, $N(v)$ is the set of neighbours for node $v$. $K$ is the total number of GNN layers, where $k=1,2,...,K$ and $H^K$ is the finalised node representations. $a{_v}^k$ is the aggregated node feature of the neighbourhood $H{_u}^{k-1}$.

\begin{equation}\label{eq-GNNeqn}
\begin{aligned}
    & a{_v}^k = Aggregate^k\{H{_u}^{k-1} : u \in N(v)\} \\
    & H{_v}^k = Combine^k\{H{_v}^{k-1}, a{_v}^k\}
\end{aligned}
\end{equation}

Graph learning includes the node-level, graph-level or edge-level tasks briefly described here.
\subsubsection*{Node-level} 
In node-level tasks, the property of each node in a graph is predicted. Node-level features may be importance-based or structure-based. 
\subsubsection*{Graph-level}
In graph-level tasks, the features of the entire graph structure are captured for computing graph similarities. Graphlet Kernel and Weisfeiler Lehman Kernel are the two main approaches for feature learning in the graph level.
\subsubsection*{Link or Edge-level}
In link-level tasks, existing links are used to make predictions on new or missing links. Distance-based, local neighbourhood overlap and global neighbourhood overlap are link-level features. 
There are four major graph learning tasks using machine learning as outlined below:
\begin{itemize}
    \item \textit{Node Classification} is the task of determining the class of a node based on classes of neighboring nodes
    \item \textit{Graph Classification} is the task of classifying an entire graph into various groups
    \item \textit{Node Regression} is the task of predicting a continuous value for a node
    \item \textit{Link Prediction} predicts potential relationships between two nodes in a graph  
\end{itemize}

 For link prediction, either node-based or subgraph-based GNN methods may be used. A GNN layer builds node-level representations, hence graph-level representations require graph pooling techniques \cite{wu2022graph} for bringing down the number of nodes in a graph. The pooling process reduces dimensionality, while retaining maximum graph information. 

 A general representation of the GNN classification pipeline is illustrated in  Fig.~\ref{fig:GNN}. Here, a GNN layer computes the input features from the input graph and aggregates these features to generate node embeddings. Node features are updated with an update function and the transformed graph is passed through a classification layer to predict labels. The most important GNN architectures used in causal learning are GCN, GAT, GGNN, GAE and GIN, which are  briefly described as follows: 

\begin{figure}
    \centering
    \includegraphics{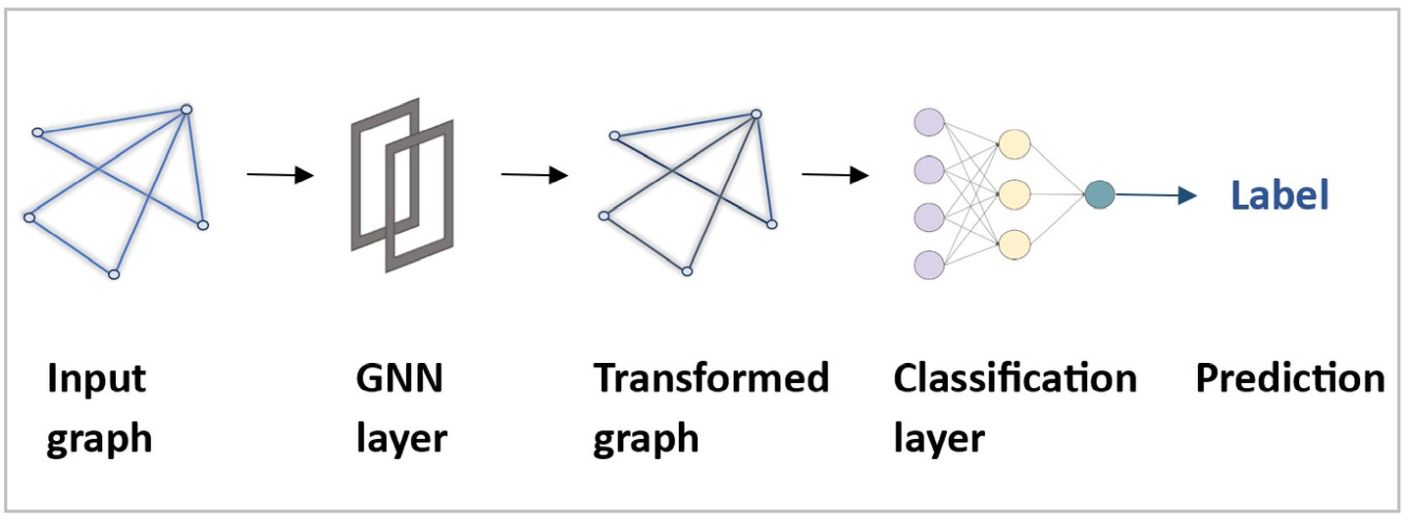}
    \caption{Prediction with Graph Neural Networks}
    \label{fig:GNN}
\end{figure}

\subsubsection{Graph Convolutional Networks (GCN)}
GCNs \cite{KipfThomasN2017SCwG} are GNNs that use convolutional aggregations and aggregate features including neighbor features. A classic GCN framework consists of graph convolutional layer(s), a linear layer and a non-linear activation layer. The input graph undergoes a convolution operation that is applied to each node, following which feature information is aggregated and the node representations updated. An activation function such as ReLU is applied to the convolutional layer output.

GraphSAGE (Graph SAmple and aggreGatE) \cite{hamilton2017inductive} is an extension of GCN with an inductive learning framework, where sampling and aggregation of features from a node's neighborhood are used to build node representations. GraphSAGE samples a subset of neighbors for each node at various layers. The aggregator then aggregates the neighbors of preceding layers with an aggregating operation using mean aggregator, pooling aggregator, or LSTM aggregator.

\subsubsection{Graph Attention Networks (GAT)}
GATs \cite{VeličkovićPetar2018GAN} incorporate an attention layer to the GNNs for compiling neighborhood properties of nodes to form embeddings. Using attention mechanism, the neighbors are assigned weights relative to their importance to a specific node. GATs attend to unseen nodes with their inductive learning, supporting both direct and indirect graphs. The graph attention layer applies a linear transformation to the node feature vectors with a weighted matrix. The activation function used is LeakyReLU and the attention coefficients are computed based on the relative importance of neighbour features. In order to have a common scaling across neighbourhood, the coefficients are normalized. The process can be improvised with multi-head attention, wherein multiple attention maps are aggregated.

\subsubsection{Gated Graph Neural Networks (GGNN)}
In GGNNs \cite{li2015gated}, Gated Recurrent Units (GRU) are incorporated with GNNs as a recurrent function for the propagation step. Recurrence is performed for a fixed number of steps and gradients are computed with backpropagation. With GGNN, node labels can be added as inputs, which are termed as node annotations. Previous hidden states and neighboring hidden states are used for updating the hidden state of a node. GGNNs need to perform the recurrent function over all nodes multiple times, which is a drawback when processing large graphs. 

\subsubsection{Graph Auto Encoder Networks (GAE)}
GAEs utilize an encoder-decoder technique to encode nodes or graphs as a latent representation to construct network embeddings. The encoder generates embeddings for nodes using graph convolutional layers. The topological information of the nodes are incorporated in the representation. The decoder reconstructs the graph adjacency matrix after evaluating pair-wise similarity of the network embeddings. The GAE aims to minimize the reconstruction loss of the decoded adjacency matrix as compared to the original matrix. 

\subsubsection{Graph Isomorphism Networks (GIN)}
GINs \cite{xu2018powerful} are GNNs with high representational power as defined by the Weisfeiler-Lehman (WL) graph isomorphism test. The WL tests if two graphs are non-isomorphic by creating subtrees for the graph nodes, followed by colour mapping each node based on the number of neighbours. The aggregate and combine functions are represented as the sum of the node features.

\section{Graph Neural Networks for Causal Learning} \label{sec-CGN}

\begin{figure*}
    \centering
    \includegraphics{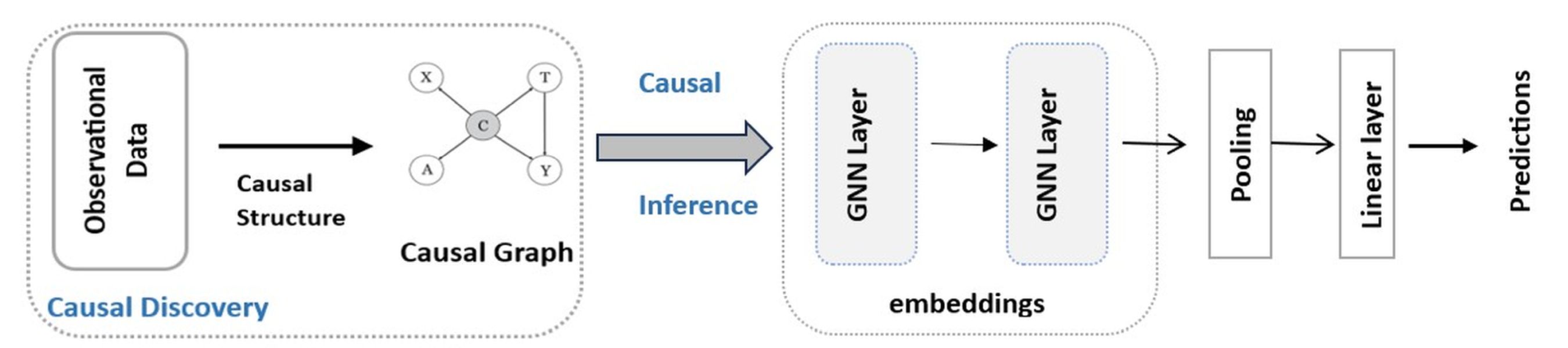}
    \caption{CLGNN Architecture}
    \label{fig:CLGNN}
\end{figure*}

\begin{figure*}
    \centering
    \includegraphics{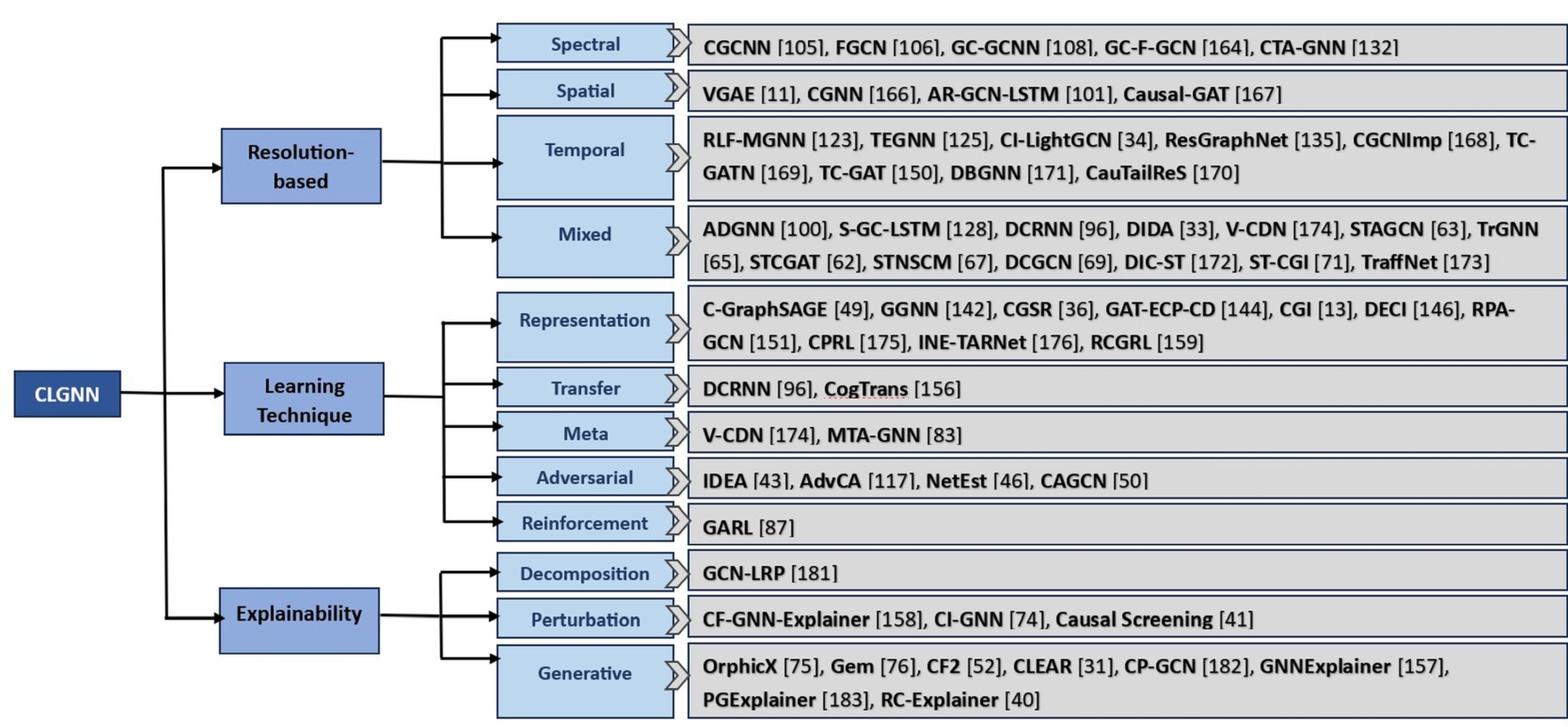}
    \caption{Taxonomy for CLGNN}
    \label{fig:CLGNNTaxonomy}
\end{figure*}

In this survey, the GNNs specifically employed for causality learning is given the term Causal Learning GNNs, abbreviated as CLGNNs and the general architecture of a CLGNN is depicted in Fig.~\ref{fig:CLGNN}. The causal discovery module infers causal structure from observational data. The causal graph is built to represent causal relationships between variables, from which causal effects are inferred using GNN. Neighborhood information in a graph is propagated through GNN layers, which is aggregated through an aggregation function. The final layer is then used to make the predictions. We group CLGNNs into three broad classes : 1) Resolution-based (pertaining to the nature of data under study), which are further categorised as Spectral, Spatial, Temporal and Mixed methods; 2) Learning Methods (determined from the learning methodologies employed for causal learning), grouped into Representation, Transfer, Meta, adversarial and Reinforcement-based method; and 3) Explainability (determined from the model explanation technique), comprised of Decomposition, Perturbation and Generative methods. A summary of datasets discussed in the survey are tabulated in Tab.~\ref{tab:tabldatasets} and Tab.~\ref{tab:tabldatasetsG}. The taxonomic structure is represented in Fig.~\ref{fig:CLGNNTaxonomy} and is discussed in this section. 

\begin{table}[hbt!]
\caption{Category-specific Datasets for CLGNNs \label{tab:tabldatasets}}
 \centering
  \begin{tabular}{|p{1.1cm}|p{4.1cm}|p{2.2cm}|}
 
  \hline
  
   \textbf{Category} & \textbf{Dataset}  & \textbf{CLGNN Studies} \\
  \hline

     \multirow{8}{*}{\shortstack{Social \\ Networks}} &  IMDB-B, IMDB-M \cite{yanardag2015deep} &  \cite{sui2022causal}, \cite{ma2022clear} \\
      
    & Yelp \cite{Yelp2022} &   \cite{zhang2022dynamic}, \cite{ding2022causal} \\
     & COLLAB \cite{yanardag2015deep}   & \cite{sui2022causal}, \cite{zhang2022dynamic}\\

      & Gowalla \cite{cho2011friendship} & \cite{ding2022causal}, \cite{wu2022causality} \\

     & Adressa \cite{gulla2017adressa} &  \cite{ding2022causal}\\

      & Pokec \cite{Takac2012-gi} & \cite{ma2021causal} \\

      & REDDIT-MULTI-5K \cite{yanardag2015deep} &  \cite{wang2022reinforced}, \cite{wang2021causal} \\
      
      & Reddit \cite{HamiltonWilliamL2018IRLo} &  \cite{tao2023idea} \\
    
      & Flickr \cite{Flickr2020} & \cite{guo2020learning}, \cite{jiang2022estimating} \\
         \hline
   
      \multirow{5}{*}{\shortstack{Citation \\ Networks}} & Cora \cite{McCallumAndrewKachites2000AtCo} & \cite{huang2022graphlime,tao2023idea,zhang2022causal,feng2021should,lee2023cagcn} \\
   
      & Citeseer \cite{10.1145/276675.276685} & \cite{tan2022learning,tao2023idea,zhang2022causal,feng2021should,lee2023cagcn} \\
  
    & Pubmed \cite{sen2008collective} & \cite{huang2022graphlime}, \cite{zhang2022causal}, \cite{feng2021should,lee2023cagcn} \\
    
     & MAG \cite{10.1145/2740908.2742839} &  \cite{cummings2020structured} \\
    
     &  ogbn-arxiv \cite{HuWeihua2021OGBD} &  \cite{tao2023idea}, \cite{feng2021should} \\
   \hline
   
  \multirow{4}{*}{\shortstack{Purchase \\ Networks}} & Transaction \cite{transaction2022} &  \cite{zhang2022dynamic} \\
    & Amazon \cite{leskovec2007dynamics}, \cite{rakesh2018linked} & \cite{ma2021causal}, \cite{wu2022causality} \\
   & ogbn-products \cite{HuWeihua2021OGBD}  & \cite{tao2023idea} \\
   & Diginetica \cite{Diginetica2016} & \cite{wu2022causality} \\
   \hline
   
    \multirow{5}{*}{\shortstack{Road \\ Networks}} & PeMS03/04/07/08 \cite{guo2021learning} &  \cite{zhao2022stcgat}, \cite{gu2022stagcn} \\
 
      & SG-TAXI \cite{sgtaxi2016} & \cite{li2021traffic} \\

       & NYC-Bike \cite{NYCBike2017} & \cite{deng2023spatio} \\
        & METR-LA \cite{jagadish2014big} & \cite{lin2023dynamic}, \cite{wang2022dynamic}, \cite{zhang2022grangerb} \\
        & PEMS-BAY \cite{li2017diffusion} & \cite{wang2022dynamic}, \cite{zhang2022grangerb} \\
       
   \hline

     \multirow{8}{*}{\shortstack{Bio- \\informatics}} & MUTAG & \cite{debnath1991structure,zheng2023ci,lin2022orphicx,lin2021generative,tan2022learning,sui2022causal,zhao2023towards} \\

   &  Mutagenicity \cite{kazius2005derivation} & \cite{wang2022reinforced}, \cite{wang2021causal} \\

   &   NCI1 \cite{wale2008comparison} & \cite{zheng2023ci,lin2022orphicx,lin2021generative,tan2022learning,sui2022causal} \\
  
   &    PROTEINS \cite{borgwardt2005protein} & \cite{zheng2023ci}, \cite{sui2022causal} \\
    & Tox21  \cite{huang2016tox21challenge}, ToxCast  \cite{Toxcast2022}  & \cite{meng2023meta} \\
       
      & SIDER \cite{kuhn2016sider}, MUV \cite{rohrer2009maximum}  &  \cite{meng2023meta} \\
    
       
  & Sachs \cite{sachs2005causal} & \cite{yang2023causal} \\
    &  Dream3 \cite{marbach2009generating} & \cite{amirinezhad2022active} \\

   \hline

 \multirow{9}{*}{\shortstack{Medicine, \\ Health}} & ABIDE \cite{di2014autism}, SRPBS \cite{tanaka2021multi}  &  \cite{zheng2023ci} \\
      
    & REST-meta-MDD \cite{yan2019reduced}  & \cite{zheng2023ci} \\
                

     & Infection \cite{faber2021comparing} &  \cite{zhao2023towards} \\

      & HCP \cite{van2013wu, hodge2016connectomedb}, UR data \cite{wein2021graph} &  \cite{wein2021graph} \\
      

      & ASIA \cite{Lauritzen1988-fd},  Cancer \cite{alma991001755139704691} &  \cite{zevcevic2021relating} \\

     & COVID-19 \cite{owidcoronavirus} & \cite{wang2022causalgnn}, \cite{ntemi2022autoregressive} \\

   & Wave1 \cite{Chantala1999-at} & \cite{ma2021causal} \\
  \hline
   \multirow{2}{*}{Psychology} & SEED \cite{duan2013differential}, SEED-IV \cite{zheng2018emotionmeter}  & \cite{kong2022causal}, \cite{li2023fusion} \\
   
      & DEAP \cite{koelstra2011deap}  & \cite{zhang2022improved} \\
         \hline

       \multirow{4}{*}{\shortstack{Advertising \\ /Sentiment}} & Criteo \cite{criteo2014}, Avazu \cite{Avazu2015}  & \cite{zhai2023causality}\\
 

     & MovieLens-1M \cite{MovieLens1M2003} & \cite{zhai2023causality}\\
      & SST \cite{socher2013recursive} &  \cite{schnake2021higher}\\
            \hline

     \multirow{4}{*}{\shortstack{Image/ \\Video}} & MNIST \cite{lecun1998gradient}, CIFAR-10 \cite{krizhevsky2009learning} & \cite{sui2022causal}\\
      & CMNIST \cite{monti2017geometric} &  \cite{sui2022adversarial} \\
   
     & DAVIS \cite{Pont-Tuset2017-nl} & \cite{varga2021fast}\\
     & ADNI \cite{ADNI2017} & \cite{tang2023causality} \\
      \hline
   
  \multirow{2}{*}{Energy} & UMass Smart \cite{UMass2017} &   \cite{wang2023short} \\

      & Energy \cite{misc_appliances_energy_prediction_374} &  \cite{xu2020multivariate} \\
        \hline

        \multirow{3}{*}{\shortstack{Trade, \\Finance}} & Nasdaq \cite{qin2017dual} &  \cite{xu2020multivariate} \\

       & UN-Comtrade \cite{UNComtrade} &  \cite{monken2021graph} \\

        & Exchangerate \cite{ExchangeRate2016} &  \cite{xu2020multivariate} \\
   
    \hline

        \multirow{2}{*}{\shortstack{Systems/ \\Industrial}} & AIOps  \cite{wang2023hierarchical} &  \cite{wang2023hierarchical} \\
         & TFF \cite{ruiz2015statistical} &  \cite{wang2023causal} \\

    \hline
     \multirow{4}{*}{\shortstack{Climate/ \\Geology}} & HadCRUT5 \cite{morice2021updated}, HadSST3 \cite{kennedy2011reassessing} &  \cite{ChenZiwei2022RGwe} \\

     &  ERSSTv4/v3b \cite{smith2008improvements}, ERA5 \cite{hersbach2020era5}, Berkeley-Earth \cite{rohde2013berkeley}  & \cite{ChenZiwei2022RGwe} \\


    

          & WADI \cite{ahmed2017wadi}, Swat \cite{mathur2016swat} &  \cite{wang2023hierarchical}\\


   \hline
    \multirow{7}{*}{\shortstack{Causal \\ datasets}}   & Causal News Corpus \cite{tan2022causal} & \cite{trust2022ggnn} \\
    & ECPE \cite{xia2019emotion} & \cite{cao2022graph} \\
    & EventStoryLine \cite{caselli2017event} & \cite{phu2021graph}\\
    & Causal-TimeBank \cite{mirza2014extracting} & \cite{phu2021graph} \\
    & SemEval-2010 task 8 \cite{hendrickx2019semeval} & \cite{gao2022joint}, \cite{yuan2023tc}, \cite{chen2022complex} \\
    & Altlex \cite{hidey2016identifying} &  \cite{chen2022complex}, \cite{yuan2023tc} \\
    & MEDCAUS \cite{moghimifar2020domain} & \cite{moghimifar2020domain} \\
      \hline
    
      \end{tabular}
  \label{tab:tabldatasets}
\end{table}

    \begin{table}[hbt!]
\caption{General Datasets for CLGNNs \label{tab:tabldatasetsG}}
 \centering
 \setlength\extrarowheight{-8pt}
  \begin{tabular}{|p{5.7cm}|p{2cm}|}
 
  \hline
    & \\
    \textbf{Dataset}  & \textbf{CLGNN Studies} \\
  \hline
    & \\

     FB15K,WN18,WN18RR \cite{bordes2013translating}, FB15K-237 \cite{toutanova2015representing} & \cite{wu2023cogtrans} \\
        
  Motif \cite{ying2019gnnexplainer} &  \cite{sui2022adversarial} \\
     Tree-Cycles \cite{ying2019gnnexplainer}  & \cite{lucic2022cf} \\
 
     Tree-Grids, BA-Shapes \cite{ying2019gnnexplainer}  & \cite{lucic2022cf}, \cite{zhao2023towards} \\
 

     Graph-SST2/SST5   \cite{yuan2022explainability} &  \cite{gao2023robust}, \cite{zhao2023towards} \\
    
     Graph-Twitter  \cite{yuan2022explainability} &   \cite{gao2023robust} \\
     Earthquake \cite{alma991001755139704691} &  \cite{zevcevic2021relating} \\

      Visual genome \cite{krishna2017visual} & \cite{wang2022reinforced}, \cite{wang2021causal} \\

     MOL-BACE \cite{wu2018moleculenet} & \cite{gao2023robust} \\

   MOL-BBBP \cite{wu2018moleculenet} & \cite{gao2023robust}, \cite{sui2022adversarial} \\
   MOL-HIV \cite{wu2018moleculenet} & \cite{sui2022adversarial}, \cite{ma2022clear}  \\

          \hline

      \end{tabular}
  \label{tab:tabldatasetsG}
\end{table}


\subsection{Resolution-based Causal Learning GNNs}\label{sec-CGNRes}
In this section, we present the CLGNNs in terms of the format of the modeled data. This aids in determining the specific GNN approaches that are capable of learning causality in the various data spaces. To this end, the resolution-based CLGNNs are further grouped into Spectral, Spatial, Temporal and Mixed classes, each of which are described in this section.
\subsubsection{Spectral}\label{sec-CGNResSpectral}
Spectral data consists of information related to frequencies obtained from signals in machinery, brain monitoring, etc. For making effective use of signal data, it is pertinent to compile the interactive information between these signals. This process is simplified by a graphical approach that facilitates the capture of the topological structure of the system \cite{wang2023causal}. The signals are converted to adjacency matrices for representing graph connections, and then transformed into causal graphs. The inference from the causal graph estimator is used for graph learning tasks. GNNs are adept at amassing global information with graph Fourier transform, wherein the graph signals are projected in the eigenvector space.

Emotion recognition has been conventionally studied using techniques such as CNNs \cite{huang2023novel}. Recent research has demonstrated the potential of GNNs for causal learning in the spectral domain \cite{kong2022causal, li2023fusion, wang2023causal}, with GCN and GAT being the most common methods. \cite{kong2022causal} employed GCNs for causal emotion recognition from multi-channel EEG signals experiments that used EEG datasets. Each node was assigned as a channel and Granger causality (GC) was calculated between each node for computing the adjacency matrix. The matrix and EEG information moved through convolution layers with Chebyshev polynomials \cite{hammond2011wavelets}, and subsequently to a depthwise separable convolution layer for extracting discriminative features for the final classification step. 

\cite{li2023fusion} employed a graph fusion strategy with GCN for emotion recognition. The approach encapsulated topological, functional, and causal features. A study on EEG and peripheral physiological signals by \cite{zhang2022improved} used GCN with GC analysis for emotion recognition. The differential entropy (DE) feature of the EEG signals formed the nodes of the graph, from which a matrix was computed to determine the edges with high causal values. GC-F-GCN \cite{zhang2022granger} is a GCN-based approach for emotion recognition that uses multi-frequency band EEG feature extraction. GC was used for computing GC matrices between EEG signals at each frequency band, with each converted to asymmetric binary GC matrices. These and DE features served as adjacency matrices and node values, respectively. The graph information from the EEG signals at different frequency bands for corresponding nodes were integrated based on GC-GCN features.

Industrial systems constitute large amounts of spectral information and are another sector where causal learning is beneficial. \cite{wang2023causal} proposed a Causal Trivial Attention GNN (CTA-GNN) that used attention mechanism and disentanglement-based causal learning with backdoor criterion for fault diagnosis. The model estimated soft masks for obtaining node and edge representations. This strategy succeeded in weakening confounding effects by avoiding shortcut features. A summary of works in Spectral CLGNN is given in Tab.~\ref{tab:tableResSpectral}.

\renewcommand{\arraystretch}{2}
    \begin{table*}[hbt!]
\caption{Spectral Causal Learning Graph Neural Networks\label{tab:tableResSpectral}}
  \centering
  \setlength\extrarowheight{-5pt}
   \begin{tabular}{|p{2.2cm}|p{0.8cm}|p{1.4cm}|p{2.6cm}|p{8.7cm}|}
 
  \hline
  & & & &  \\
  \textbf{Approach}  & \textbf{Method}  & \textbf{Domain} & \textbf{Problem} & \textbf{Key Functionality} \\

    \hline
     & & & &  \\
    CGCNN \cite{kong2022causal} & GCN & Psychology & Emotion Classification &  Emotion Recognition from EEG signals with Granger Causality (GC)  \\

      FGCN \cite{li2023fusion} & GCN & Psychology & Emotion Classification  & EEG-based emotion recognition using GCN with graph fusion strategy \\

       GC-GCNN \cite{zhang2022improved} & GCN & Psychology & Emotion Classification & EEG-based emotion recognition using convolutional GNNs with GC  \\
    
    GC-F-GCN \cite{zhang2022granger} & GCN & Psychology & Emotion recognition & GCN-based EEG emotion recognition with GC \\

      CTA-GNN \cite{wang2023causal} & GAT & Industrial & Fault diagnosis & GAT with soft mask estimation and causal learning using disentanglement \\
    \hline

      \end{tabular}
  \label{tab:tableResSpectral}
\end{table*}


\subsubsection{Spatial}\label{sec-CGNResSpatial}
In Spatial data, observations are spatially associated  with each other. GCN is the most commonly used type of GNN employed in the spatial domain, where the convolution operation is carried out on each node and the weights are shared across all locations. The node features are aggregated in layers by a permutation-invariant function and the information is amassed within a localised boundary \cite{DeyuBo2023ASoS}.

\cite{ma2021causal} studied causal effects under network interference and employed Hilbert-Schmidt Independence Criterion (HSIC) for a feature space dependence test. \cite{zevcevic2021relating} also adopted an interventional approach for causal density and causal effect estimation with autoencoding. The GNN and VGAE model detected interventional change including successive changes. \cite{li2023causality} employed  multiple-instance learning with GNNs for capturing spatial proximity and the similarity of features for cancer detection using Contrastive Mechanism to segregate non-causal features. 

GNN is effective in capturing spatially adjacent nodes in a graph and was efficiently implemented by \cite{varga2021fast} with label propagation for interactive video object segmentation using a Watershed algorithm. The graph was created from superpixel segments with dimensionality reduction and causality estimated based on optical flows. Causal-GAT with disentangled causal attention (DC-Attention) was proposed by \cite{liu2023causal} for fault detection by incorporating disentangled representations with GAT. The causal graph was structured using monitoring variables, and DC-Attention was used to generate node representations from causal relations. \cite{tang2023causality} proposed a causality aware GCN framework for rigidity assessment in neuro-diseases. The GCN was built on node, structure, and representation levels, and incorporated causal feature selection. A non-causal perturbation strategy with an invariance constraint ascertained the validity of the model under varying distributions. An autoregressive (AR) GCN framework with LSTM called AR-GCN-LSTM was developed by \cite{ntemi2022autoregressive} for epidemic case prediction. A GC adjacency matrix formed the input to GCN to extract causal information of locations that influenced each other's case numbers. AR modeling in AR-GCN-LSTM assisted in capturing the linear dependencies from time series, and feature representation was extracted with two LSTM layers. A summary of works related to CLGNNs in the spatial domain is given in Tab.~\ref{tab:tableResSpatial}.

\renewcommand{\arraystretch}{2}
\begin{table*}[hbt!]
\caption{Spatial Causal Learning Graph Neural Networks\label{tab:tableResSpatial}}
  \centering
\setlength\extrarowheight{-5pt}
  \begin{tabular}{|p{2.6cm}|p{0.8cm}|p{1.1cm}|p{2.6cm}|p{8.7cm}|}
  \hline
 & & & &  \\
   \textbf{Approach}  & \textbf{Method}  & \textbf{Domain} & \textbf{Problem} & \textbf{Key Functionality} \\

    \hline
     & & & &  \\
     GNN, HSIC \cite{ma2021causal} & GNN & Multiple & Network interference & Causal estimation with intervention policy optimization for network data \\
    
   GNN, VGAE \cite{zevcevic2021relating} & GNN & Multiple & Causal Inference & Autoencoded intervention for causal density and causal effect estimation \\

    CGNN \cite{li2023causality} & GNN & Medicine & Cancer detection & Multiple-instance learning to study proximity and similarity of tumors \\

    GCN \cite{tang2023causality} & GCN & Medicine & Neurology assessment & Causality aware GCN for rigidity assessment in patients with neurodisease \\

         AR-GCN-LSTM \cite{ntemi2022autoregressive} & GCN & Medicine & Case prediction & Autoregressive GCN with LSTM for Covid-19 case prediction \\

    GNN \cite{varga2021fast} & GNN & Video & Object segmentation & GNN with label propagation for interactive video object segmentation  \\
 
    Causal-GAT \cite{liu2023causal} & GAT & Industry & Fault detection &  Causal GAT with disentangled representations for fault detection\\
    \hline
      \end{tabular}
  \label{tab:tableResSpatial}
\end{table*}

\subsubsection{Temporal}\label{sec-CGNResTemporal}
Temporal data consists of observations related to time or date instances. \cite{wang2023short} proposed an LSTM-based GNN model referred to as RLF-MGNN, for load forecasting. Multiple temporal correlations of energy usage across households were captured along with transfer entropy to build a causality graph. The collective influence of household energy usage was employed for forecasting, with only positive correlations considered in building graphs.

A transfer entropy-based causal analysis (TEGNN) was used in an experiment by \cite{xu2020multivariate} for multivariate time series (MTS) forecasting, where causal prior information was captured using the transfer entropy matrix. Time series forecasting using GraphSAGE with residual neural networks (ResGraphNet) was studied by \cite{ChenZiwei2022RGwe} for the causal prediction of global monthly mean temperature. Using the embedded ResNet, the connection weights of each node were used to build a learning framework. The training time for ResGraphNet was much higher than that of a GraphSAGE base model. \cite{wang2023hierarchical} studied hierarchical GNNs for temporal root cause analysis in systems monitoring. The model integrated topological and individual causal discoveries by incorporating intra-level and inter-level system relationships, while disregarding system logs. \cite{ding2022causal} proposed a GCN-based model called CI-LightGCN for retraining recommender systems. Causal incremental graph convolution (IGC) and colliding effect distillation were employed, with IGC performing aggregation of only new neighbors. GCN was also employed for MTS imputation by \cite{liu2022cgcnimp}, who proposed CGCNImp that incorporated GC alongside correlation and temporal dependencies. Attention mechanism and total variation reconstruction were used for retrieving latent temporal information.
  
A GAT framework called TC-GAT was proposed by \cite{yuan2023tc} for temporal causality discovery from text. TC-GAT integrated temporal aspects with causality using graph attention by computing joint weights of temporal and causal features. A causal knowledge graph was used to obtain an adjacency matrix towards causality extraction. GAT was also proposed by \cite{li2022tc} for industrial MTS forecasting, incorporating GC. Nonlinear interaction of node features was performed with parallel GRU encoders in the graph neighborhood space, and then the encoder hidden states were aggregated with attention mechanism. In this approach, large scale graphs introduce nonlinear interactive patterns that increase complexity.

 GGNNs is another causal learning approach proposed by \cite{zeyu2023causal}, who employed a framework called CauTailReS  session recommendation with counterfactual reasoning. Popularity bias was eliminated using deconfounded training with causal intervention and do-calculus. GGNN was used to generate node embeddings, and both user interest and consistency embeddings were captured for learning causality. GNNs were also employed with De Bruijn graphs using DBGNN framework to learn causal topological patterns in dynamic graphs by \cite{qarkaxhija2022bruijn}. De Bruijn graphs were used to incorporate non-Markovian characteristics of causal walks, leading to a causality-aware node classification process. A summary of works related to CLGNNs in the temporal domain is given in Tab.~\ref{tab:tableResTemporal}.

\renewcommand{\arraystretch}{2}
\begin{table*}[hbt!]
\caption{Temporal Causal Learning Graph Neural Networks\label{tab:tableResTemporal}}
  \centering
  \setlength\extrarowheight{-5pt}
     \begin{tabular}{|p{2.8cm}|p{1.2cm}|p{1.4cm}|p{2.3cm}|p{8cm}|}
  \hline
 & & & &  \\
   \textbf{Approach}  & \textbf{Method}  & \textbf{Domain} & \textbf{Problem} & \textbf{Key Functionality} \\
  \hline
   & & & &  \\
       RLF-MGNN \cite{wang2023short} & GNN & Energy & Load Forecasting & Forecasting with LSTM and transfer entropy  \\

     TEGNN \cite{xu2020multivariate} &  GNN & Multiple & MTS forecasting  & MTS forecasting with transfer entropy-based causal analysis \\

     Hierarchical GNN \cite{wang2023hierarchical}&  GNN & Systems & Rootcause Analysis & Systems causal analysis for identifying root cause of system issues \\

     CI-LightGCN \cite{ding2022causal} &  GCN & Recommender & System retraining & Recommender retraining with causal incremental graph convolution \\

    ResGraphNet\cite{ChenZiwei2022RGwe} &  GraphSAGE & Climate & TS forecasting & Global monthly mean temperature prediction using GraphSAGE \\
 
  CGCNImp \cite{liu2022cgcnimp} &  GCN & Multiple & TS imputation & Imputation with GCN integrating correlation and temporal dependency\\

  TC-GATN \cite{li2022tc} &  GAT & Industry & MTS forecasting & Multivariate time series (MTS) forecasting with GAT, GC and GRU \\
    
    TC-GAT \cite{yuan2023tc} &  GAT & Text & Causality discovery & Causality extraction with GAT using temporal and causal relations \\

    DBGNN \cite{qarkaxhija2022bruijn} &  GNN & Multiple & Causality discovery & GNN with De Bruijn graphs to learn causal patterns in dynamic graphs \\
 
    CauTailReS \cite{zeyu2023causal} &  GGNN & Recommender & Rrecommendation & Session recommendation with counterfactual reasoning using GGNN \\
    \hline
  \end{tabular}
  \label{tab:tableResTemporal}
\end{table*}

\subsubsection{Mixed}\label{sec-CGNResMixed}
In this section, we discuss causal graphs that handle mixed resolutions such as spatial-temporal data. In spatio-temporal graphs, node connections are formed as a function of time and space, and are applicable in instances where both time and location/space are of value in causal learning. Spatio-temporal CLGNNs have been beneficial for epidemic studies\cite{wang2022causalgnn}, economic \cite{monken2021graph} and traffic \cite{gu2022stagcn, li2021traffic, zhao2022stcgat, deng2023spatio} forecasting etc. 

\cite{wang2022causalgnn} proposed ADGNN, an attention-based dynamic GNN with causal learning for forecasting epidemic cases. The study included causal features and constraints with a minimal use of parameters. Many traffic flow prediction studies incorporated attention mechanism with GNN for causal learning, with both \cite{gu2022stagcn} and \cite{zhao2022stcgat} developing frameworks for capturing causal traffic dynamics using attention-based GNNs. The former proposed STAGCN, which considered both global and local traffic dynamics, assuming no interaction between the static and dynamic graph units. The latter work proposed the STCGAT framework, which captured overall spatio-temporal dependencies using local and global causal convolutions. \cite{li2021traffic} developed TrGNN, a trajectory-based model for predicting traffic flows using graphs that encapsulated temporal dependencies using attention mechanism. In a similar domain, \cite{deng2023spatio} proposed STNSCM for building a causal model with Frontdoor criterion for handling confounding bias in bike flow prediction. The framework integrated counterfactual reasoning and demonstrated effective long term predictions. \cite{lin2023dynamic} proposed Dynamic Causal GCN (DCGCN) for modeling spatio-temporal dependencies in the traffic prediction domain. Time-varying dynamic causal graphs were incorporated for constructing superior spatio-temporal topology representation. A similar approach was proposed by \cite{wang2022dynamic} for spatio-temporal forecasting with dynamic causality analysis. The causality adjacency matrix was fed to GCN for extracting dynamic correlations in road networks. 

GCN was also used for spatio-temporal cellular traffic forecasting by \cite{zhang2022dic} to develop a framework integrating GCN, referred to as DIC-ST. This framework was based on decomposition and integration with causal structure learning. An empirical mode decomposition (EMD) method was adopted for multi-scale decomposition, with integration of various time series analysis of three different components. Following EMD, integration was performed based on KNN to subsequently employ the integration of prediction results. ST-CGI \cite{zhang2022grangerb} employed GC with an auto-regressive process to develop an interpretable traffic prediction model. Dilated causal convolution network was employed for encoding temporal information, with causal relationships extracted from the embeddings using GNNs. \cite{xu2023traffnet} proposed TraffNet for causal learning of traffic volumes through path embedding, route learning, and road segment embedding. Bi-GRU was employed for path embedding, following which, a meta-path based GAT was used for route learning to determine the origin-destination demands for each route. Road segment embedding was obtained by aggregating all path embeddings using a position-aware message passing technique for capturing the road segment position. Finally, these embeddings were fed to a temporal GRU module for forecasting.

GAT is also useful in causal analysis of social and purchase networks that demand exploring both spatial and temporal elements. Disentangled Intervention-based Dynamic (DIDA) GAT proposed by \cite{zhang2022dynamic} handles distribution shifts in dynamic graphs, using a disentangled attention-enabled causal inference framework to capture invariant patterns in graphs. For economic forecasting , \cite{monken2021graph} proposed S-GC-LSTM for trade flow causal analysis of outlier events using Stateless Graph Convolutional LSTM, to capture events that affect trading. A counterfactual study was performed, mainly focusing on highly disruptive events across geographical areas. 

Spatio-temporal networks are also relevant in neuroscience, where brain functions need to be analysed anatomically over time. DCRNN model proposed by \cite{wein2021graph} used GNNs with  convolutional and recurrent networks for inference of causal relations in brain networks. Causal structural-functional interactions in brain regions were identified with the diffusion-convolution approach. Spatial and time factors also play a role in predicting future movements in visual intelligence as researched by \cite{li2020causal}. These authors proposed a V-CDN model with GNN as a spatial encoder, supplemented with counterfactual predictions. A GCN-based model employed by \cite{cummings2020structured} for predicting papers with high future citation counts, used percentile thresholds for ranking. A summary of works related to CLGNNs in the mixed domain is given in Tab.~\ref{tab:tableResMixed}.

\renewcommand{\arraystretch}{2}
\begin{table*}[hbt!]

\caption{Mixed Resolution Causal Learning Graph Neural Networks\label{tab:tableResMixed}}
  \centering
  \setlength\extrarowheight{-5pt}
   \begin{tabular}{|p{2.2cm}|p{0.8cm}|p{1.4cm}|p{2.6cm}|p{8.7cm}|}
  \hline
 & & & &  \\
   \textbf{Approach}  & \textbf{Method}  & \textbf{Domain} & \textbf{Problem} & \textbf{Key Functionality} \\
  \hline
   & & & &  \\
    ADGNN  \cite{wang2022causalgnn} & GNN & Health & Epidemic Forecasting & Attention-based dynamic GNN for forecasting COVID-19 cases \\

    S-GC-LSTM \cite{monken2021graph} & GCN & Trade & Economic Forecasting & Trade flow causal analysis of outlier events with Graph Convolutional LSTM \\

       DCRNN \cite{wein2021graph} & GCN & Medicine & Causal neuroscience & Causal analysis of brain networks with convolutional and recurrent networks  \\

     DIDA \cite{zhang2022dynamic} & GAT & Multiple & Link Prediction & Disentangled attention-enabled approach to get invariant patterns in graphs \\
    
      V-CDN \cite{li2020causal} & GNN & Visual Intel. & Prediction & Predict future movements with GNN as spatial encoder \\

      STAGCN \cite{gu2022stagcn} & GCN & Urban Intel.  & Traffic Forecasting & Traffic trends are captured with an attention-based GCN \\

     TrGNN \cite{li2021traffic} & GAT & Urban Intel.  & Traffic flow prediction & Trajectory-based model for predicting traffic flows with GAT \\

     STCGAT \cite{zhao2022stcgat} & GAT & Urban Intel.  & Traffic flow prediction & Causality based GAT with local and global causal convolutions for prediction \\
    
     STNSCM \cite{deng2023spatio} & GCN & Urban Intel.  & Bike flow prediction & Causal model with frontdoor criterion for handling confounding bias \\
  
    DCGCN \cite{lin2023dynamic} & GCN & Urban Intel.  & Traffic prediction & Dynamic Causal GCN for modelling spatiotemporal dependencies \\

    GCN \cite{wang2022dynamic} & GCN & Urban Intel.  & Traffic prediction & Dynamic Causal GCN for spatio-temporal forecasting in traffic systems \\
    
    DIC-ST \cite{zhang2022dic} & GCN & Urban Intel. & Traffic prediction & Framework for spatio-temporal forecasting with decomposition, integration \\
    
     ST-CGI  \cite{zhang2022grangerb} & GCN & Urban Intel. & Traffic prediction & Causal inference and autoregressive processes for interpretable prediction \\
    
        TraffNet \cite{xu2023traffnet} & GAT & Urban Intel. & Traffic prediction &  Causal learning with path embedding, route learning, segment embedding \\

    GCN \cite{cummings2020structured} & GCN & Citation & Node Classification & Predict academic papers with high citation counts in future \\
    \hline

  \end{tabular}
  \label{tab:tableResMixed}
\end{table*}
\subsection{Causal Learning GNNs based on learning methods}\label{sec-CGNLearn}

In this section, we discuss CLGNNs employed in five learning areas, such as Representation, Transfer, Meta, Adversarial and Reinforcement learning. The works related to CLGNNs in the different learning paradigms are summarised in Tab.~\ref{tab:tableLearning}.

\renewcommand{\arraystretch}{2}
\begin{table*}[hbt!]
\caption{Causal Learning Graph Neural Networks- Learning Methods\label{tab:tableLearning}}
  \centering
\setlength\extrarowheight{-5pt}
   \begin{tabular}{|p{.4cm}|p{2.3cm}|p{1.8cm}|p{1.8cm}|p{9.5cm}|}
  \hline

 
  \hline
   & & & &  \\
 \textbf{Type} &  \textbf{Approach} & \textbf{Graph Method} & \textbf{Domain} & \textbf{Key Functionality} \\

  \hline
   & & & &  \\
   \parbox[t]{2mm}{\multirow{14}{*}{\rotatebox[origin=c]{90}{\textbf{Representation}}}} & GAT, CAL  \cite{sui2022causal} & GAT & Multiple & Attention-based GNN with mitigation of confounding effects for graph classification \\

  & GraphFwFM \cite{zhai2023causality} & Graph-SAGE & Advertising & Causality-based CTR prediction with integrated representation learning \\

    & C-GraphSAGE \cite{zhang2022causal} & Graph-SAGE & Citation &  GraphSAGE with causal sampling for reducing perturbation influence \\
    & GGNN \cite{trust2022ggnn} & GGNN & Event causality & Causal event detection with GGNN to capture semantic and syntactic information \\
    & CGSR \cite{wu2022causality} & GNN & Recommendation & Session-based recommendation with causality and correlation graph modeling \\
    & GAT-ECP-CD \cite{cao2022graph} & GAT & Text & Textual emotion-cause pair causal relationship detection with GAT \\
    & CGI \cite{feng2021should} & GCN & Citation & Causal GCN Inference model studying causal effects of node local structure  \\
    & GCN \cite{guo2020learning} & GCN & Social Network & Network deconfounder for learning representations to uncover hidden confounders \\
    & DECI  \cite{phu2021graph} & GCN & ECI & Document level event causality identification with GCN \\
    & GCN \cite{gao2022joint} & GCN & ECI & Extract event causality with GCN using textual, knowledge enhancement channels \\
    & RPA-GCN \cite{chen2022complex} & GCN & Text & Head-to-tail entity annotation approach for text with GAT and GCN \\
     & CPRL \cite{yang2022causal} & GCN & Text & Causal pattern representation learning using BERT, GCN and attention mechanism \\
     & INE-TARNet \cite{adhikari2023inferring} & GNN & General & Causality in presence of heterogeneous peer influence using GNN-based estimator \\
    & RCGRL \cite{gao2023robust} & GNN & General & Learn representations against confounding effects with Instrumental Variable  \\
    \hline

 & & & &  \\
      \parbox[t]{2mm}{\multirow{2}{*} {\rotatebox[origin=c]{90}{\textbf{TL}}}} &  DCRNN \cite{wein2021graph} & GNN & Medicine & Transfer Learning enabled inference of causal relations in brain networks  \\

      & CogTrans \cite{wu2023cogtrans} & GCN & KGR & Attention-based cognitive transfer learning for Knowledge Graph Reasoning   \\
    \hline

     & & & &  \\
     \parbox[t]{2mm}{\multirow{2}{*} {\rotatebox[origin=c]{90}{\textbf{Meta}}}} & V-CDN \cite{li2020causal} & GNN & Visual & Long term predictions from video sequences based on causal relationships \\

    & MTA, GNN \cite{meng2023meta} & GNN &  Drug discovery &  Meta learning with motif-based task augmentation (MTA) technique \\
    \hline

 & & & &  \\
 \parbox[t]{2mm}{\multirow{5}{*} {\rotatebox[origin=c]{90}{\textbf{Adversarial}}}} & IDEA \cite{tao2023idea} & GNN & Multiple  & Defense on graph attacks by developing invariance objectives from causal features \\
    & AdvCA  \cite{sui2022adversarial} & GNN & Multiple &  Graph augmentation strategy to address covariate shift in OOD generalization \\
        & NetEst \cite{jiang2022estimating} & GNN & Social network & Causal inference using GNN for learning representations for confounders \\
    & GCN \cite{moghimifar2020domain} & GCN & Text & Adaptive causality identification and localisation with GCN \\
    & CAGCN \cite{lee2023cagcn} & GCN & Citation & Causal attention GCN with node and neighbour attention for robust learning \\
    \hline

 & & & &  \\
    \parbox[t]{2mm}{\multirow{2}{*} {\rotatebox[origin=c]{90}{\textbf{RL}}}} & GNN \cite{amirinezhad2022active} & GNN & General & A GNN-based model with Q-iteration to extract causality with minimal intervention \\
    
    & GARL \cite{yang2023causal} & GAT & General & RL for causal discovery with graph attention for embedding structure information\\

    \hline
    
  \end{tabular}
  \label{tab:tableLearning}
\end{table*}

\subsubsection{Representation Learning}\label{sec-CGNRepLearn}
Graph representation learning involves learning a model from graph-structured data by building features or embeddings that represent the structure. GraphSAGE is the most commonly used GNN approach for inductive representation learning, and was integrated with causal inference to develop a framework called C-GraphSAGE \cite{zhang2022causal} for classification. C-GraphSAGE incorporated causal sampling with the purpose of controlling the influence of perturbations. C-GraphSAGE performed well under perturbation conditions, although GAT was superior with no perturbation. In \cite{trust2022ggnn}, causal relationships between events in social-political news were extracted with GGNNS. The authors modeled event causality identification with contextualized language representations by building a graph representation of all documents in the Causal News Corpus dataset, along with their dependencies. GGNN layers were stacked to form an encoder, with information aggregated based on the edge type and direction, and GRU applied for node embedding updates. 

\cite{sui2022causal} proposed Causal Attention Learning (CAL) with mitigation of confounding effects using softmask estimation from attention scores. The graph is decomposed to causal and trivial attended graphs with two GNN layers. The authors proposed disentanglement of causal and trivial features, with GNNs filtering shortcut patterns for capturing causal features. An attention-based framework called Causality and Correlation Graph Modeling for Effective and Explainable Session-based Recommendation (CGSR) was proposed by \cite{wu2022causality} for session-based recommendation with causality and correlation graph modeling. CGSR has four components viz. graph construction, item representation learner, session representation learner and recommendation score generator. Effect graph, cause graph and correlation graph were constructed, following which, a weighted GAT was used for item representation learning on each of these graphs. The session representation learner forms a session representation by aggregating learned item representations in the session sequence using an attention layer. \cite{cao2022graph} also proposed a GAT-based method for causality detection referred to as GAT-ECP-CD for textual emotion-cause pair causality. The study used BiLSTM to obtain semantic representation. The sentence vector was fed to GAT for capturing dependencies between clauses, followed by a joint prediction layer with three stages of predictions.

\cite{zhai2023causality} developed a structured causality-based representation learning approach called GraphFwFM, incorporating a Field-weighted Factorization Machines (FWFM) mechanism for CTR prediction, with the GNN-based graph representation learning consolidating feature graphs, user graphs, and ad graphs. \cite{gao2023robust} developed a Robust Causal Graph Representation Learning (RCGRL) framework to learn representations against confounding effects for improving causality prediction and generalisation performance. The variant of the IV approach was used in the framework, where conditional moment restrictions for confounding elimination were transferred to unconditional restrictions. Causal GCN Inference (CGI) was proposed by \cite{feng2021should} for studying the causal effects of the local structure of a node, when labels of neighboring nodes vary in GCN. As a first step, intervention was done on predictions by blocking the graph structure, followed by a comparison with original predictions for determining the causal effects of the local structure. A GCN-based network deconfounder was proposed by \cite{guo2020learning} for learning representations to uncover hidden confounders from network information. The representation learning function was parameterized using GCN towards learning causal effects. These authors customised the BlogCatalog dataset studied by \cite{li2019adaptive} through a synthesis of outcomes and treatments. 

A GCN-based framework called DECI was proposed by \cite{phu2021graph} for document level event causality identification (ECI) using interaction graphs. The interaction graph nodes were formed from all the words, event mentions, and entity mentions in a document. Node connections were formed from discourse-based, syntax-based, and semantic-based information. The interaction graphs and representation vectors were regularized for improved representation learning. \cite{gao2022joint} also proposed an approach for event causality extraction using GCN. The model was enhanced using a dual-channel approach with textual (TEC) and knowledge (KEC) enhancement channels. TEC learns significant intra-event features and KEC uses GCN for assimilating external causality transition knowledge. GCN and GAT were employed by \cite{chen2022complex} to develop relation position and attention-GCN (RPA-GCM) for establishing complex causal relations in text by marking entity boundaries. Relation features were extracted by integrating an attention network with a dependency tree. The interaction information between entities and relations were captured using a bi-directional GCN. GCN was employed for developing Causal Pattern Representation Learning (CPRL) \cite{yang2022causal} for extracting causality from text using an entity set of risk factors for various diseases. Input encoding was performed by BERT, followed by GCN for encoding dependencies, and causal features were extracted with an attention mechanism for constructing causal pattern representation. \cite{adhikari2023inferring} devised INE-TARNet for estimating causal effects in the presence of heterogeneous peer influence using a GNN-based estimator. Arbitrary assumptions regarding network structure, interference conditions, and causal dependence were captured by the model. These assumptions were encoded by Network Structural Causal Model (NSCM), which generated Network Abstract Ground Graph (NAGG) for reasoning about treatment effects under arbitrary network interventions. Further, NSCM and NAGG were employed for extracting individual network effects (INE) using GNN.

\subsubsection{Transfer Learning}\label{sec-CGNTransferLearn}

Transfer learning (TL) is a technique in which a machine learning model trained on a specific task is re-purposed to a related problem in a similar domain. Research on transfer learning with GNN is limited and experiments so far demonstrate the transfer process to be plausible when source and target graphs are alike. This drawback, referred to as negative transfer, impacts target performance \cite{kooverjee2022investigating}. 

A diffusion convolution recurrent neural network (DCRNN) based on GNN was employed by \cite{wein2021graph} to infer causal relations in brain networks. Their model was pre-trained on large volumes of fMRI sessions, ensuring better results for small datasets on account of diffusion-convolution used on massive graphs leading to memory errors. Moreover, the authors experimented on the model using TL, with the results indicating a considerable reduction in error rates. \cite{wu2023cogtrans} proposed CogTrans for cognitive transfer learning-based Knowledge Graph Reasoning (KGR) using GCN, where transferring was based on hierarchical structure similarity. The approach retained causal structure, with hierarchical random walk being employed to acquire entity and relation characteristics for pre-training in the knowledge graph. Multi-head self attention mechanism was used for the reasoning phase in which an attention score is aggregated from entity and relation layers for encoding.


\subsubsection{Meta Learning}\label{sec-CGNMetaLearn}

In machine learning, meta learning (ML) serves as a solution for data scarcity by exploiting previously learned experiences towards learning an algorithm that generalizes across various tasks. Model-Agnostic Meta-Learning (MAML) \cite{finn2017model} is the most commonly adopted approach for training GNNs \cite{mandal2022metalearning} and trains model parameters for accelerated learning with minimal gradient updates. \cite{meng2023meta} proposed ML with a motif-based task augmentation (MTA) technique for molecule property prediction, targeted at addressing the few-shot learning challenge associated with this task. Causal substructure was studied, with experiments conducted on molecule datasets. The model recovers the most relevant motifs from a previously defined motif vocabulary for label generation, for the purpose of task augmentation. Once augmented, classification was performed based on euclidean distances. \cite{li2020causal} experimented on custom video datasets to develop a ML framework called V-CDN for predicting future visuals. In V-CDN, Perception and Inference modules draw a graph inference, and the dynamics module was used to predict future visuals from the inferred graph. Inference was extracted from causal relationships and the graph distribution was inferred based on the image representation for prediction.


\subsubsection{Adversarial Learning}\label{sec-CGNAdvLearn}

Adversarial Learning (AL) is a machine learning technique that examines and devises defenses against adversarial attacks on models. Based on adversarial capability, attacks may be of the poisoning or evasion kinds. Poisoning attacks involve training time attacks that affect networks such as GNN through data poisoning. Evasion attacks are test time attacks that add poisoned data at test time. Defense models for graph data serve to stabilise the model in the course of adversarial events.

\cite{tao2023idea} proposed a framework called Invariant causal DEfense method against adversarial Attacks (IDEA) to develop defense methods against adversarial attacks on GNNs. The defense was devised by developing invariance objectives from causal features and utilized domain partitioner to manage mixed domains. IDEA handled evasion and poisoning attacks, with invariant defending plausible only with linear causality. A graph augmentation strategy called AdvCA (Adversarial Causal Augmentation) proposed by \cite{sui2022adversarial} addresses the problem of covariate shift in Out-of-distribution (OOD) generalization. AdvCA performed adversarial data augmentation while preserving causal feature across various environments. The covariate shift was devised based on the domain using graph size and type or colour. \cite{jiang2022estimating} proposed Networked causal effects Estimation (NetEst) for inferring causality in network settings. The model used GNN for learning representations for confounders. Additionally, two AL modules were introduced for allowing mismatched distributions to follow uniform distributions based on embeddings from confounders. NetEst presented a framework for addressing the issue of distribution gaps through representations. \cite{moghimifar2020domain} proposed a GCN-based domain adaptive causality identification and localisation with AL for extracting causal relationships in text. Distribution shifts were minimised using a gradient reversal approach. AL was applied to training domain discriminator for differentiating the source and domains. Moreover, feature representation was trained to overcast the domain discriminator. CAGCN by \cite{lee2023cagcn} employed causal attention GCN with node and neighbour attention for enhanced robustness. AL was performed based on nettack (graph structure and attributes are attacked), metattack (global attack using meta learning) and random attack (random addition of edge to graph).

\subsubsection{Reinforcement Learning (RL)}\label{sec-CGNReinfLearn}

\begin{figure}
    \centering
    \includegraphics{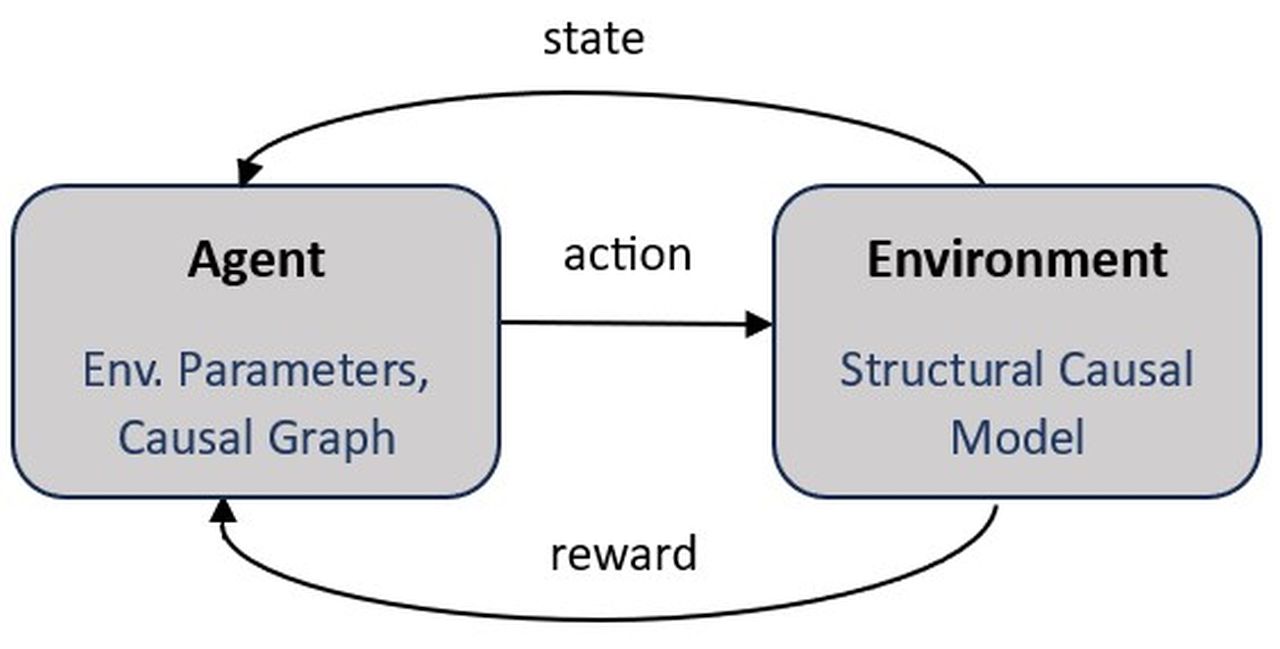}
    \caption{Causal Reinforcement Learning}
    \label{fig:CRL}
\end{figure}

Causal reinforcement learning explores causal mechanisms for agent learning process towards improved decision making. The agent acts in the environment and the effects of actions are observed followed by counterfactual analysis. A general structure of the causal reinforcement learning process is illustrated in Fig.~\ref{fig:CRL}. The environment and the agent are connected through the causal model and causal graph. Here, the action may be observational, interventional, or counterfactual.

\cite{amirinezhad2022active} researched on learning causal structures from interventional data using GNN and RL. An active learning approach was employed, where the intervention results were used to derive causal relationships for formulating causal structures with minimal interventions. Training was performed using a Q-iteration algorithm and the completed partially DAG (CPDAG) for each sub-network was derived from the causal network to serve as an input to another network, whose output in turn was used for intervention. A graph attention RL (GARL) framework was proposed by \cite{yang2023causal} for causal discovery, which employed a GAT for embedding structural information and prior knowledge along with RL to generate variable orderings.


\subsection{Causal Learning GNNs based on Explainability}\label{sec-CGNExplain}

Explainability of a machine learning model refers to the extent to which a model's output is meaningful. The terms explainability and interpretability are commonly interchanged, though interpretability focuses on the cause and effect from a model. With reference to GNN, interpretability is inherently designed in the GNN architecture and explanations are post-hoc. This review details explainability with reference to methods adopted for causal GNNs. The explainability techniques applied to CLGNNs are summarised in Tab.~\ref{tab:tableExplain}.

\renewcommand{\arraystretch}{2}
\begin{table*}[hbt!]
\caption{Explainable Causal Graph Neural Networks \label{tab:tableExplain}}
  \centering
  \setlength\extrarowheight{-5pt}
   \begin{tabular}{|p{1.6cm}|p{3cm}|p{0.9cm}|p{1.5cm}|p{8.9cm}|}

  \hline
   & & & &  \\
  \textbf{Method} &  \textbf{Approach} & \textbf{Graph} & \textbf{Domain} & \textbf{Key Functionality} \\


     \hline
      & & & &  \\
{\multirow{1}{*}{\textbf{Decomposition}}} & GCN, LRP \cite{holzinger2021towards} & GCN & Medicine & Explainable multi-modal causability framework for medical decision modeling \\

     \hline
 & & & &  \\
   {\multirow{3}{*} {\textbf{Perturbation}}} & CF-GNN-Explainer \cite{lucic2022cf} & GNN & General & Counterfactual explainer for GNN  with minimal perturbation  \\
     & CI-GNN \cite{zheng2023ci} & GNN & Medicine & Explainable causal GNN with VAE to extract brain functionality connections \\
   
    
    & Causal Screening \cite{wang2021causal} & GNN & General & Explainable GNN by selecting a graph feature with large causal attribution \\
     
  \hline

   & & & &  \\
    {\multirow{8}{*} {\textbf{Generative}}} & OrphicX \cite{lin2022orphicx} & VGAE & Bioinformatics & Explanations from latent causal factors using information flow mechanism \\

    & Gem \cite{lin2021generative} & GNN & Bioinformatics & Provides auto-encoder based explanations with Granger Causality \\

    & CF2  \cite{tan2022learning} & GNN & Multiple & Explain node classification with counterfactual and factual reasoning \\
    & CLEAR \cite{ma2022clear} & VGAE & Multiple & Graph classification with counterfactual explanation \\
     & CP-GCN \cite{jin2022supporting} & GCN & Text & Relation extraction from medical text with causality-pruned dependency forest \\
  & GNNExplainer \cite{ying2019gnnexplainer} & GNN & Multiple & Explainable graph classification using mask generation \cite{zhao2023towards} \\
  & PGExplainer \cite{luo2020parameterized} & GNN & Multiple & Explainable graph classification using mask generation \cite{zhao2023towards} \\
    & RC-Explainer \cite{wang2022reinforced} & GNN & Multiple & Explainability in a sequential decision process uncovering edge attributions   \\

    \hline

  \end{tabular}
  \label{tab:tableExplain}
\end{table*}

The standard methods used for GNN explanations are surrogate, decomposition, perturbation, generative, and gradient methods \cite{yuan2022explainability}. Surrogate models for graphs are more challenging to implement, owing to their topological design. Though surrogate approaches such as GraphLime and PGM-Explainer are used as graph explainers, ambiguity exists in defining the neighbors in the input graph. Similarly, gradient methods have been proven to be less advanced in GNN explanation compared to other explainers \cite{lin2021generative}, \cite{wang2021causal}. Hence, with a view of causal application, only decomposition, perturbation and generative approaches are discussed in this study.


\subsubsection{Decomposition methods}\label{sec-CGNExplainDecomposition}

Decomposition methods decompose the original network into various elements for determining the most significant features constituting the importance scores. Using score distribution rules, prediction scores are distributed from the output layer with a backpropagation approach towards the input layer. The importance scores for node features account for edge importance, as well as walk importance. Layer-wise Relevance Propagation (LRP) is a decomposition method that follows the law of total probability. \cite{schnake2021higher} extended LRP to develop a technique adapted to GNNs referred to as GNN-LRP, which form the score decomposition rule with high-order Taylor decomposition. A key difference from LRP is that GNN-LRP distributes scores to graph walks and not nodes or edges \cite{yuan2022explainability}. \cite{holzinger2021towards} presented GNN-LRP with GCN for developing an explainable multi-modal causability framework for information fusion across various feature spaces. A joint multi-modal representation was proposed to be computed in a decentralized manner towards model scalability and security. This approach was proposed as an exploration-based explainable method with counterfactual graphs for building an automated decision model.


\subsubsection{Perturbation methods}\label{sec-CGNExplainPerturbation}

Perturbation methods investigate variations in output on changing inputs to a model, whereby the changes in output specify the input elements relevant for inference. GNNExplainer, PGExplainer, SubgraphX, Causal Screening, and CF-GNNExplainer are a few of the perturbation-based explainer methods used for explaining graph models, with the latter two specifically designed for causality. CF-GNNExplainer, a counterfactual explainer for GNNs, proposed by \cite{lucic2022cf} used minimal perturbation to the input graph for varying predictions. The framework incorporated edge deletions for generating counterfactual explanations. \cite{wang2021causal} developed causal screening for explainable GNN for classification, by selecting a graph feature with large causal attribution. The approach involved identifying predominant edges for explainable classification. CI-GNN, another explainable GNN with a focus on causality, was developed by \cite{zheng2023ci} to extract functionality connections in brain networks using GraphVAE to learn disentangled latent representations.

\subsubsection{Generative methods}\label{sec-CGNExplainGenerative}

In the generative approach, graph generation methods are employed to provide explanations by devising a search strategy to obtain the most explanatory subgraph. GNN generative explainable models include approaches such as mask generation, VGAE, GAN, Diffusion, and RL \cite{ying54generative}. CLEAR \cite{ma2022clear} is a VGAE approach for optimising graph data for generalization on unseen graphs with counterfactual explanation generation. A GCN was used as the encoder and the decoder was a MLP, which also learned the mean and covariance of the latent variables' prior distribution. The input graph and counterfactuals were matched using graph matching. OrphicX \cite{lin2022orphicx} is also a VGAE approach that provided explanations from latent causal factors using an information flow mechanism and backdoor adjustment. In the model, linear independence of explained features was not assumed. Gem, another VGAE approach proposed by \cite{lin2021generative}, provided explanations with GC. A rapid explanation process with no requirement of prior knowledge of GNN structure was employed. The framework received the input graph into the explainer to output a compact explanation graph without prior structural knowledge of the GNN. \cite{tan2022learning} developed a framework called CF2 for explaining node classification with counterfactual and factual reasoning, with evaluation done based on the probability of necessity and sufficiency. RL-based GNN explanation called RC-Explainer was proposed by \cite{wang2022reinforced}, where the explainable model was defined in a sequential decision making process that uncovered edge attributions for causal screening. In the framework, edge dependencies were considered for deriving causal effects.

Similar to Gem, PGExplainer also provided local and global views for explanations with parameterized networks. A distinction from the Gem framework is that PGExplainer, a mask generation method, learns MLP from the target GNN's node embeddings \cite{lin2021generative}. PGExplainer learned discrete edge masks by training a parameterized mask predictor. Edge embeddings were obtained from an input graph by combining node embeddings. These edge embeddings are used for predictions, following which, the discrete masks were sampled with reparameterization. The mask predictor training was performed by assimilating the original and model predictions. GNNExplainer, which is also a mask generation method, learned both edge and node soft masks for explaining predictions. Similar to PGExplainer, the masks are combined with the original graph, and the original and new predictions were assimilated. \cite{zhao2023towards} extended PGExplainer and GNNEXplainer with a three-layer GCN by aligning internal embeddings of the raw graph and the explainable subgraph. A three-layer GCN was employed in these two approaches for evaluating their faithfulness and consistency in comparison to other explanatory approaches such as GRAD, Gem, and RG-Explainer. The proposed approach, mainly PGExplainer, demonstrated commendable performance compared to other explainers, along with the base PGExplainer and GNNEXplainer. \cite{jin2022supporting} proposed Causality-Pruned semantic dependency forest Graph Convolutional Network (CP-GCN ) for relation extraction from text using causality-pruned dependency forests. Semantic and syntactic information was introduced to the dependency tree for constructing forests. A task-specific causal explainer was then trained for pruning dependency forests, which was subsequently fed to GCN for learning representations.


\section{Challenges and Directions}\label{sec-CD}

There exists several challenges associated with causal learning and GNNs, which we discuss in this section.

\textbf{Data Quality:} Observational data, often marred by incompleteness and biases such as selection bias and confounding, presents a unique set of challenges in the context of causal learning. Traditional modeling assumptions that address these challenges might not always encapsulate the system's intricate dynamics. For instance, deriving causality from observational data can be hindered by issues such as deficient anomaly detection approaches and the complexity of certain treatments, such as those related to images.

\textit{Opportunities:} Constructed data, derived from scenarios similar to observational studies, emerges as a promising solution, especially when ground truth is elusive. This approach aids in evaluating causality and can be complemented by weak supervision for causal feature selection in cases with sparse labeled data. Furthermore, the integration of interventional and observational data can enhance causal learning. However, there remains challenges such as dataset imbalances and discrepancies in training and test set distributions \cite{castro2020causality}. Future endeavours should prioritize exploring methods to address these challenges, such as techniques to handle dataset shifts caused by factors like irregular data collection. Delving into distributional disparities, aspects like population, annotation, and manifestation shifts, along with considerations for anticausal tasks, need thorough investigation. Additionally, the development of models that can seamlessly adapt to these shifts and biases will be pivotal in advancing the field of causal learning.

\textbf{Causal Assumptions:} Causal discovery makes strong assumptions such as Markov property, and causal inference makes assumptions such as SUTVA, consistency etc. Both require ongoing investigation in scenarios that deviate from these premises. Similarly, assumptions such as positivity are not adhered to, when certain elements in a system remain idle or untouched. Furthermore, causal studies generally represent treatments as discrete events, which is inadequate for exploring causality in continuous data.

\textit{Opportunities:} Sensitivity of causal study outcomes to violations of these assumptions must be investigated to design a flexible framework for validating causal models. A concrete validation approach needs to be designed for validating causal estimates in the presence of potential confounding, leading to unbiased estimators. Estimating causal effects in a continuous environment requires further research to model treatments as continuous events. 

\textbf{Research scalability:} Many causal datasets are small-scale datasets, which are typically scaled-down graph datasets that do not align with real-world dynamics. Synthetic or semi-synthetic datasets are the core of many causal experiments. 

\textit{Opportunities:} Translating new approaches in the context of high dimensional data needs to be further investigated using larger causal data. This is particularly significant, since the primary step in preparing high-dimensional data for causality learning is encoding. This is adequate for non-semantic data, however, the viability of encoding semantic data to a low dimensional representation is largely domain-dependent. Low dimensional features must encompass treatment, outcomes, and confounding information, to lead to representation learning. Few areas requiring investigation in representation learning are adequate management of confounding information and noisy outcomes.

\textbf{Dynamic causality:} 

Causality research is largely based on static observational data studies. Causal inference for modeling dynamic data would pave the way for capturing causality in dynamic environments. Moreover, causal inference in the spatial domain requires the inclusion of spatial heterogeneity, spatial interactions, and spatial lag effects, for accurate causal estimation. This will open a path for designing dedicated spatial causal learning methods \cite{akbari2023spatial}.

\textit{Opportunities:} If the mechanisms associated with causal structure varies to the extent of affecting causal links, distribution shift must be incorporated in modeling causality \cite{glymour2019review}. Causality-based feature selection can handle distribution shift data, but this requires interventional information for extracting causal variables. Moreover, it is challenging to capture causal dynamics in a continuous environment, on account of the difficulty in gathering time series data rapidly enough to account for the swift changes occurring in the system. Additionally, for spatial causal analysis, development of appropriate tools competent enough to capture spatial complexities would pave the way for more significant research in the domain.

\textbf{Multiple tasks and multi-modal causality:} There is a dearth of studies exploring causal learning with multi-modal data. Publicly available multi-modal causal datasets can motivate research in this area and enhance causal applicability in multiple fields. Causal representations for multiple tasks can be extracted by clustering trajectories from the various domains, though determining the granularity of causal variables is largely dependent on the individual task. 

\textit{Opportunities:} Diligent curation of multi-modal datasets would uplift multi-modal research to some degree. Additionally, taking into consideration the impracticality of validating causal models with several tasks and interventions, meta learning needs to be further researched alongside reinforcement learning in the field of causality.

\textbf{Training data:} While GNNs are adept at learning complex patterns in data, their effectiveness is largely dependent on the volume of training data. GNNs reach high performance and generalizability with large datasets, which presently have limited availability. Furthermore, such studies with limited data tend to result in biased outcomes. Conversely, when large-scale data is available, experiments are often restricted to data subsets on account of the computational requirements of GNNs. 

\textit{Opportunities:} Transfer learning can be utilized in circumstances where data is limited. This approach is also beneficial for learning networks with constant changes. Both instances are made possible by model transferability through the process of training a model with data-rich historical or topical subsets \cite{jiang2022graph}. In scenarios with large-scale data, graph partitioning is one possible approach, in which networks are split into smaller units \cite{jiang2022graph}. Generative adversarial networks (GAN) can also enhance training data for GNN models with a generator-discriminator approach, with Wasserstein GAN (WGAN) \cite{arjovsky2017wasserstein} able to generate data for graph modeling purposes.

\textbf{Graph structures and Scalability:} Graph modeling in a multi-tasking framework still remains a challenge, with problems such as forecasting in various task areas demanding multiple graph structures. Most studies address this problem by employing GNN on the same graph using feed-forward layers, leading to multiple results \cite{jiang2022graph}. Furthermore, scalability is an open problem in processing massive graphs. Development of GNN architectures that are scalable for continually evolving graphs is critical for speeding up the graph training process. 

\textit{Opportunities:} Neighbor sampling and mini-batch processing are a few approaches currently used by researchers \cite{hamilton2017inductive}, but require further study towards overcoming the overfitting and oversmoothing problems of GNNs. Meta learning is another approach that can equip GNNs with the ability to learn on a task for generalising to multiple tasks. When a model learned on a graph is applied to other graph structures, interactions between multiple graphs are not captured. This includes incorporation of cross-network node similarities, anchor links, etc. Similarly, graph correlations are also not accounted for, hence, information from multiple graphs must be consolidated in representation learning for effective modeling.

\textbf{Complexity of graph structure and graph layers:} GNN models are mostly studied on homogeneous graphs on account of the complexity of implementing GNNs as heterogeneous structures. This is particularly important for capturing dynamic spatial information. Moreover, this aspect also limits GNN's capability in handling multi-modal data. Multi-modal GNN research primarily uses balanced data, thus requires further exploration in the case of imbalanced datasets. Besides, adding unlimited layers in GNN models can contribute to a performance decline \cite{zhang2019graph}, making this a possible research direction for developing deep robust GNNs that are computationally efficient. 

\textit{Opportunities:} Modeling in complex environments requires research into higher-order structures such as graphlets. Laplacian matrix estimator and data adaptive graph generation are some existing approaches proposed to capture dependencies in a dynamic environment \cite{jiang2022graph}. Incorporating embedding propagation operations is a strategy adopted in this area. Further research into dynamic and heterogeneous graph-structures would contribute to stability and adaptability of GNNs in varying structures. Furthermore, the challenge associated with increasing GNN depth can be alleviated to some degree by adopting skip connection based structures \cite{zhang2019graph}, however, more flexible architectures are required for building deeper models. In complex networks, nodes and edge attributes may be dynamic with a multilayered structure. Each layer must be defined as a separate dimension, necessitating encoding at these various network layers. Studies must be expanded on encoding dynamic networks incorporating node attributes for predicting topology dynamics. 

\textbf{Adversarial attacks on graphs:} A major drawback of the GNN architecture is that it is susceptible to adversarial attacks, leading to performance decline. A benchmark approach for building defenses against adversarial attacks on GNNs is required to develop a suitable framework for defense against graph attacks. The framework must incorporate structure as well as attribute perturbations for building robust GNNs. In addition to formulating defense approaches, more research should be conducted on identifying and cleaning contaminated graphs. 

\textit{Opportunities:} Measuring perturbations on graphs are undetectable at human level and hence require robust perturbation evaluation measures to address this problem. Adversarial attacks on graphs are mostly researched on static graphs with node attributes \cite{jin2021adversarial}. An important research direction in adversarial studies is a focus on complex graphs with edge attributes, as well as dynamic graphs. Transferability of graph adversarial examples is another area that requires further research for building efficient and robust graph models.

\textbf{Explainability of graph data:} The complex nature of graph data leads to abstract explanations, which can be further compounded by a lack of domain knowledge. This problem can be alleviated to some extent by developing standardised datasets for explanation tasks. Moreover, heterogeneous data contributes to complex structures, which can complicate the explainability process. Most explainable GNN models are instance-level and not at the model level. Explainable GNNs are critical for identifying the subgraphs that  more significantly contribute towards model outcomes. Lack of locality information and varying node neighbors make explaining graph structures a challenging task. Extending existing explainable methods to GNNs is not very dependable on account of the topology information in adjacency matrices being presented as discrete values. For the same reason, input optimization approaches commonly utilised in explaining models cannot be extended to GNNs. Learning softmasks is another approach used as an explainability technique, but if translated to an adjacency matrix, this will interfere with its discrete nature \cite{yuan2022explainability}. 

\textit{Opportunities:} Explaining GNNs involves understanding structural information of graphs, which is not directly explainable by current methods. Evaluation of explainable methods in the graph domain is not straightforward, since graph visualisations are not easy to render for direct human understanding \cite{li2022explainability}. Both task-specific and task-agnostic evaluation metrics must be researched and developed with a view towards quantitative explanation of GNN models.


\section{Conclusion} \label{sec-CON}

This survey delivered a thorough examination of the state-of-the-art GNNs in the domain of causal learning.  In recent years, there has been a surge in the use of GNN-based methods for causal learning, which has revolutionized the field. Drawing from our thorough research of peer-reviewed publications, it is evident that these GNN-based methods have amplified the efficiency, resilience, and versatility of causal learning. Modern techniques are markedly more adaptable compared to their traditional counterparts, and they harness a more extensive data spectrum during the initial processing stages. This monumental shift necessitates a novel taxonomic approach, and thus, we have grouped the methods into three distinct categories: Resolution-based, Learning-based, and Explainability-based. We delved into the advancements within each category, highlighting the unique contributions and trajectories for each of these approaches. Additionally, we provided a concise guide of the resources frequently cited in the literature, including both datasets and the overall framework of resources used in researching CLGNNs. To conclude, we have identified several challenges and open research avenues that warrant attention in this evolving field.

\section*{Acknowledgments}
This work is supported by the SAGE Athena Swan Scholarship.

\FloatBarrier
\small
\bibliographystyle{IEEEtranN}
\bibliography{gnnsurvey}

\end{document}